# Inducing Interpretable Voting Classifiers without Trading Accuracy for Simplicity: Theoretical Results, Approximation Algorithms, and Experiments

**Richard Nock** RNOCK@MARTINIQUE.UNIV-AG.FR
*Université Antilles-Guyane*
GRIMAAG-*Département Scientifique Interfacultaire*
*Campus Universitaire de Schoelcher*
*B.P. 7209*
*97275 Schoelcher, Martinique, France*

## Abstract

Recent advances in the study of voting classification algorithms have brought empirical and theoretical results clearly showing the discrimination power of ensemble classifiers. It has been previously argued that the search of this classification power in the design of the algorithms has marginalized the need to obtain interpretable classifiers. Therefore, the question of whether one might have to dispense with interpretability in order to keep classification strength is being raised in a growing number of machine learning or data mining papers. The purpose of this paper is to study both theoretically and empirically the problem. First, we provide numerous results giving insight into the hardness of the simplicity-accuracy tradeoff for voting classifiers. Then we provide an efficient "top-down and prune" induction heuristic, WIDC, mainly derived from recent results on the weak learning and boosting frameworks. It is to our knowledge the first attempt to build a voting classifier as a base formula using the weak learning framework (the one which was previously highly successful for decision tree induction), and not the strong learning framework (as usual for such classifiers with boosting-like approaches). While it uses a well-known induction scheme previously successful in other classes of concept representations, thus making it easy to implement and compare, WIDC also relies on recent or new results we give about particular cases of boosting known as partition boosting and ranking loss boosting. Experimental results on thirty-one domains, most of which readily available, tend to display the ability of WIDC to produce small, accurate, and interpretable decision committees.

## 1. Introduction

Recent advances in the study of voting classification algorithms have brought empirical and theoretical results clearly showing the discrimination power of ensemble classifiers (Bauer & Kohavi, 1999; Breiman, 1996; Dietterich, 2000; Opitz & Maclin, 1999; Schapire & Singer, 1998). These methods basically rely on voting the decision of individual classifiers inside an ensemble. It is widely accepted, and formally proven in certain cases (Schapire, Freund, Bartlett, & Lee, 1998; Schapire & Singer, 1998), that their power actually relies on the ability to build potentially *very* large classifiers. It has even been observed experimentally that such an ensemble can sometimes be as large as (or larger than) the data used to build the ensemble (Margineantu & Dietterich, 1997) ! Then, a simple question arises, namely what is the interest a customer can have in using such a classifier, instead of simple lookups in the data, and using algorithms such as nearest neighbor classifiers (Margineantu & Dietterich, 1997) ?

After some of the most remarkable recent studies in voting classification algorithms, some authors have pointed out the interest to bring this classification power to data mining, and more precisely to make interpretability a clear issue in voting classification algorithms (Bauer & Kohavi, 1999; Ridgeway, Madigan, Richardson, & O'Kane, 1998). Some authors go even further, and argue that the importance of interpretability has been marginalized in the design of these algorithms, and put behind the need to devise classifiers with strong classification power (Ridgeway et al., 1998). But interpretability also governs the quality of a model by providing answers to how it is working, and, most importantly, why. According to Bauer & Kohavi (1999), striving for comprehensibility in voting models is one of the principal problems requiring future investigations. They also remark that "voting techniques usually result in incomprehensible classifiers that cannot easily be shown to users".

Comprehensibility is, on the other hand, a hard mining issue (Buja & Lee, 2001) : it depends on parameters such as the type of classifiers used, the algorithm inducing the classifiers, the user mining the outputs, etc. . Though the quantification of interpretability is still opened in the general case (Buja & Lee, 2001), there are some clues coming from theory and practice of machine learning/data mining indicating some potentially interesting requirements and compromises to devise an efficient learning/mining algorithm.

A first requirement for the algorithm is obviously its generalization abilities: without classification strength, it is pointless to search for interesting models of the data. A second requirement, more related to mining, is the size of the classifiers (Nock & Gascuel, 1995; Nock & Jappy, 1998). If accurate, a classifier with restricted size can lead to faster and deeper understanding. This is obviously not an absolute rule, rather an approximate proxy for interpretability : pathologic cases exist in which, for example, a large and unbalanced tree can be very simple to understand (Buja & Lee, 2001). Note that in this example, the authors explain that the tree is simple because all its nodes can be described using few *clauses*. Therefore, simplicity is also associated to a short description, but using a particular class of concept representation.

A third parameter influencing comprehensibility is the nature of the algorithm's output. Inside the broad scope of symbolic classifiers, some classes of concept representations appear to offer a greater comfort for interpretation. Decision trees belong to this set (Breiman, Freidman, Olshen, & Stone, 1984), though they also raise some interpretability problems : Kohavi & Sommerfield (1998) quote that

> "the clients [business users] found some interesting patterns in the decision trees, but they did not feel the structure was natural for them. They were looking for those two or three attributes and values (*e.g.* a combination of geographic and industries) where something "interesting" was happening. In addition, they felt it was too limiting that the nodes in a decision tree represent rules that all start with the same attributes."

Although not limiting from a classification viewpoint, the ordering of nodes prior to classification can therefore make it uncomfortable to mine a decision tree. Notice that this problem might hold for any class of concept representation integrating an ordering prior to classification: decision lists (Rivest, 1987), alternating decision trees (Freund & Mason, 1999), branching programs (Mansour & McAllester, 2000), etc. . There exists, however, a type of classifiers on which related papers appear to be generally unanimous on their mining abilities : disjunctive normal form formulas (DNFs, and their numerous extensions), that is, disjunctions of conjunctions. Interestingly enough, this is the class which motivated early works (and a great amount of works afterwards) on the well-known PAC theory of learning (Valiant, 1984, 1985), partly because of the tendency humans seem to have to represent knowledge using similarly shaped rules (Valiant, 1985). This class is also the dual of the one implicitly used by Buja & Lee (2001) to cast their size measure for decision trees (to state whether the concept represented is simple or not).

It is our aim in this paper to propose theoretical results and approximation algorithms related to the induction of very particular voting classifiers, drawing their roots on simple rule sets (like DNF), with the objective to keep a tradeoff between simplicity and accuracy. Our aim is also to prove that, in the numerous induction algorithms already proposed throughout the machine learning and data mining communities, some of them, previously used in decision trees and decision lists induction, can be easily adapted to cope with this objective, thereby leading to easy-to-implement (and compare) algorithms. The next section presents a synthesis of our contribution, which is detailed in the rest of the paper.

## 2. Our Contribution

This paper is principally concerned with the theoretical and experimental study of a set of voting classifiers which we think is likely to provide an accurate answer to the simplicity-accuracy tradeoff: decision committees (DC) (Nock & Gascuel, 1995). DC is informally the Boolean multiclass extension of polynomial discriminant functions. A decision committee contains rules, each of these being a pair (monomial, vector). Each monomial is a condition that, when fired, returns its vector. After each monomial has been tested, the sum of the returned vectors is used to take the decision. This additive fashion for combining rules is absent from classical Boolean classifiers such as Decision Trees (DT) or Decision Lists (DL). Furthermore, unlike these two latter classes, the classifier contains absolutely no ordering, neither on variables (unlike DT), nor on monomials (unlike DL). When sufficiently small DCs are built and adequate restrictions are taken, a new dimension in interpreting the classifier is obtained, which does not exist for DT or DL. Namely, any example can satisfy more than one rule, and a DC can therefore be interpreted by means of various rule *subsets* (in a naive conversion of a DT or a DL into rule sets, any example satisfies exactly one rule). Decision committees resemble or generalize other rule sets (Cohen & Singer, 1999). In this paper, the authors consider DNF-shaped formulas, in which the output of a monomial is not a class (called “positive”), but a (non-negative) confidence in the classification as positive. A default class predicts the other class, called “negative” (this is a setting with two classes). Computing the class of an observation boils down to summing the confidences of the rules it satisfies, and then deciding the positive class if the sum is greater than zero, and the negative class otherwise. Decision committees are a generalization of these formulas, in which we remove the setting’s constraint (two classes) and authorize the membership prediction to arbitrary classes, thereby leading to a true voting classifier. This voting fashion is a feature that decision committees share with decision tables (Kohavi & Sommerfield, 1998). However, decision tables classifiers are based on majority voting of the examples (and not of rules), over a restricted "window" of the description variables. They necessitate the storing of many examples, and the interpretations of the data can only be made through this window, according to this potentially large set of examples. Decision committees rather represent an efficient way to encode a voting method into a small number of rules, and the way a class is given can be brought back to early works in machine learning (Clark & Boswell, 1991). More formal details are provided in the next section.

Among our theoretical results, that are presented in the following section, we provide formal proofs that the simplicity-accuracy tradeoff is also hard to achieve for DC, as well as for the construction of complex votes involving DT. This last result shows that, while mixing C4.5 with boosting provides one of the most powerful classification algorithms (Friedman, Hastie, & Tibshirani, 2000), pruning boosting is essentially heuristic (Margineantu & Dietterich, 1997).

The algorithm we propose for the induction of DC, WIDC (for Weak Induction of Decision Committees), has the following key features. It uses recent results on partition boosting, ranking loss boosting (Schapire & Singer, 1998) and some about pruning Boolean formulas (Kearns & Mansour, 1998). WIDC follows a scheme close to C4.5's for decision trees (Quinlan, 1994), or ICDL's for decision lists (Nock & Jappy, 1998) ; as such, it differs from previous studies in voting classifiers (boosting, bagging (Breiman, 1996)) by features such as the fact that no modification is made on the example's distribution during induction. It is also one if its differences with the SLIPPER rule induction approach (Cohen & Singer, 1999).

On multiclass and multilabel problems, WIDC proposes a very fast and simple solution to ranking loss boosting, optimal in fairly general cases, and asymptotically optimal in most of the remaining ones. The general problem of ranking loss boosting was previously conjectured $NP$-Hard (Schapire & Singer, 1998). Though our ranking loss boosting algorithm is not always optimal, we also show that the general ranking loss boosting problem related to Schapire & Singer (1998) is actually not $NP$-Hard, and can be solved in polynomial time, though it seems to require the use of complex and time-expensive algorithms, related to the minimization of (symmetric) submodular functions. This also partially justifies the use of our simple and fast approximation algorithm.

The last section of this paper presents experimental results obtained with WIDC on thirty-one domains, most of which are readily available and can be found on the UCI repository of machine learning database (Blake, Keogh, & Merz, 1998).

In order to keep the paper self-contained and as concise and readable as possible, we have chosen to put an appendix at the end of the paper containing all proofs of our results.

## 3. Decision Committees

Let $c$ be the number of classes. Unless otherwise specified, an example $e$ is a couple $e = (o, c_o)$ where $o$ is an observation described over $n$ variables, and $c_o$ its corresponding class among $\{0, 1, ..., c-1\}$ ; to each example $(o, c_o)$ is associated a weight $w((o, c_o))$, representing its appearance probability with respect to a learning sample $LS$ which we dispose of. $LS$ is itself a subset of a whole domain which we denote $\mathcal{X}$. Obviously, we do not have entire access to $\mathcal{X}$ $(LS \subset \mathcal{X})$ : in general, we even have $|LS| \ll |\mathcal{X}|$ ($|.|$ denotes the cardinality; we suppose in all that follows that $\mathcal{X}$ is discrete with finite cardinality). In the particular case

where $c = 2$, the two classes are noted "-" ($c_o = 0$) and "+" ($c_o = 1$), and called respectively the negative and positive class. The learning sample is the union of two samples, noted $LS^-$ and $LS^+$, containing respectively the negative and positive examples. It is worthwhile to think the positive examples as belonging to a subset of $\mathcal{X}$ containing all possible positive examples, usually called the *target concept*.

As part of our goal in machine learning, is the need to build a reliable approximation to the true classification of the examples in $\mathcal{X}$, that is, a good approximation of the target concept, by using only the examples in $LS$. Good approximations shall have a high accuracy over $\mathcal{X}$, although we do not have access to this quantity, but rather to its estimator: a more or less reliable accuracy computable over $LS$. We refer the reader to standard machine learning books (Mitchell, 1997) for further considerations about this issue. A DC contains two parts:

- A set of unordered pairs (or rules) $\{(t_i, \vec{v}_i)\}_{i=1,2,...}$ where each $t_i$ is a monomial (a conjunction of literals) over $\{x_1, \overline{x}_1, x_2, \overline{x}_2, ..., x_n, \overline{x}_n\}^n$ ($n$ being the number of description variables, each $x_j$ is a positive literal and each $\overline{x}_j$ is a negative literal), and each $\vec{v}_i$ is a vector in $\mathbb{R}^c$. For the sake of readability, this vectorial notation shall be kept throughout all the paper, even for problems with only two classes. One might choose to add a single real rather than a 2-component vector in that case.

- A Default Vector $\vec{D}$ in $[0,1]^c$. Again, in the two-class case, it is sufficient to replace $\vec{D}$ by a default class in $\{+, -\}$.

For any observation $o$ and any monomial $t_i$, the proposition "$o$ satisfies $t_i$" is denoted by $o \Rightarrow t_i$. The opposite proposition "$o$ does not satisfy $t_i$" is denoted by "$o \not\Rightarrow t_i$". The classification of any observation $o$ is made in the following way: define $\vec{V}_o$ as follows

$$\vec{V}_o = \sum_{\substack{(t_i, \vec{v}_i) \\ o \Rightarrow t_i}} \vec{v}_i \ .$$

The class assigned to $o$ is then:

- $\arg\max_j \vec{V}_o$ if $|\arg\max_j \vec{V}_o| = 1$, and

- $\arg\max_{j \in \arg\max_{j'} \vec{V}_o} \vec{D}$ otherwise.

In other words, if the maximal component of $\vec{V}_o$ is unique, then the index gives the class assigned to $o$. Otherwise, we take the index of the maximal component of $\vec{D}$ corresponding to the maximal component of $\vec{V}_o$ (ties are solved by a random choice among the maximal components).

DC contains a subclass which is among the largest classes of Boolean formulas to be PAC-learnable (Nock & Gascuel, 1995), however this class is less interesting from a practical viewpoint since rules can be numerous and hard to interpret. Nevertheless, a subclass of DC (Nock & Gascuel, 1995) presents an interesting compromise between representational power and interpretability power. In this class, which is used by WIDC, each of the vector components are restricted to $\{-1, 0, +1\}$ and each monomial is present at most once.

The values $-1$, $0$, $+1$ allow natural interpretations of the rules, being either in favor of the corresponding class ($+1$), neutral with respect to the class ($0$), or in disfavor of the corresponding class ($-1$). This subclass, to which we relate as $\mathrm{DC}_{\{-1,0,+1\}}$, is, as we now prove, suffering the same algorithmic drawbacks as DT (Hyafil & Rivest, 1976) and DL (Nock & Jappy, 1998): even without restricting the components of the vectors, or with any restriction to a set containing at least one real value, the construction of small formulas with sufficiently high accuracy is hard. This is a clear motivation for using heuristics in decision committee's induction.

## 4. Building Small Accurate Decision Committees (and Alike) is Hard

We now show that building decision committees is a hard algorithmic task when one strives to obtain both small and accurate formulas. There are two usual notions of size which can naturally be used for decision committees. The first one is the whole number of literals of the formula (if a literal is present $i$ times, it is counted $i$ times) (Nock & Gascuel, 1995; Nock & Jappy, 1998), the second one is the number of rules of the formula (Kearns, Li, Pitt, & Valiant, 1987). Our results imply that regardless of the restriction over the values of the vectors (as long as they are elements of a set with cardinality $\geq 2$), and already for two-classes problems, minimizing the size of a decision committee for both size definitions is as hard as solving well-known $NP$-Hard problems. Therefore, the task is also hard for $\mathrm{DC}_{\{-1,0,+1\}}$ with the particular values $-1$, $0$, $+1$ for the vectors.

### 4.1 The Size of a DC is Measured as its Whole Number of Literals

**Theorem 1** *When the size of a DC is measured as its whole number of literals, it is $NP$-Hard to find the smallest decision committee consistent with a set of examples $LS$.*

**Proof:** See the Appendix. □

We can easily adapt Theorem 1 to the case where the rules are replaced by weighted DT as advocated in boosted C4.5 (Schapire & Singer, 1998). Here, each tree returns a class $\in \{+1, -1\}$, and each tree is given a real weight to leverage its vote. The sign of the linear combination gives the class of an example. The following theorem holds again with any limitations on the leveraging coefficients (as long as at least one non-zero value is authorized), or without limitation on the coefficients. By this, we mean that for each of the applicable limitations (or without), the problem is $NP$-Hard. The size notion is the sum, over all trees, of their number of nodes.

**Theorem 2** *It is $NP$-Hard to find the smallest weighted linear combinations of DT consistent with a set of examples $LS$, without limitation on the leveraging coefficients, or for any possible limitation, as long as at least one non-zero value is authorized.*

While it is well known that boosting results in a rapid decreasing of the error over $LS$ which can easily and rapidly drop down to zero (as long as it is possible), Theorem 2 shows that attempts to efficiently reduce the size of the vote when boosting DT is $NP$-Hard. If the problem is simplified to the to pruning of a large consistent vote of DT (Margineantu & Dietterich, 1997), to obtain a smaller consistent (or with limited error) vote with restricted size, it is again possible (using the same reduction) to show that this brings $NP$-Hardness.

### 4.2 The Size of a DC is Measured as its Number of Rules

We now state and prove the equivalent of Theorem 1 with this new size notion.

**Theorem 3** *When the size of a DC is measured as its number of rules, it is $NP$-Hard to find the smallest decision committee consistent with a set of examples $LS$. The result holds even when the concept labeling the examples is a monotone-DNF formula, that is, a disjunction of conjunction (DNF), each without negative literals.*

**Proof:** See the Appendix. □

A previous work (Kearns et al., 1987) proves a similar theorem concerning the minimization of the size of a DNF. Theorem 3 can be shown to be more general, as the class of $\text{DC}_{\{-1,0,+1\}}$ with two rules strictly contains that of DNF with two monomials.

The statement of Theorems 1, 2, 3 as optimization problems was chosen for pure convenience ; replacing them by their associated decision problems (decide whether there exist a consistent formula whose size is no more than some fixed threshold) would trivially make the problems not only $NP$-Hard, but also $NP$-Complete.

## 5. Overview of WIDC

An algorithm, IDC, was previously proposed (Nock & Gascuel, 1995) for building decision committees. It proceeds in two stages. The first stage builds a potentially large subset of different rules, each of which is actually a $\text{DC}_{\{-1,0,+1\}}$ with only one rule. In a second stage, it gradually clusters the decision committees, using the property that the union of two $\text{DC}_{\{-1,0,+1\}}$s with different rules is still a $\text{DC}_{\{-1,0,+1\}}$. At the end of this procedure, the user obtains a set of DCs, and the most accurate one is chosen and returned. Experimental results display the ability of IDC to build small DCs. In that paper, we provide an algorithm for learning decision committees which has a different structure since it builds only one DC. More precisely, WIDC is a three stage algorithm. It first builds a set of rules derived from results on boosting decision trees (Schapire & Singer, 1998). It then calculates the vectors using a scheme derived from Ranking loss boosting (Schapire & Singer, 1998). It finally prunes the final $\text{DC}_{\{-1,0,+1\}}$ using two possible schemes: a natural pruning which we call "pessimistic pruning", and pruning using local convergence results (Kearns & Mansour, 1998), which we call "optimistic pruning". The default vector is always chosen to be the observed distribution of ambiguously classified examples.

### 5.1 Building a Large Decision Committee using Partition Boosting

Suppose that the hypothesis (not necessarily a decision committee, it might be *e.g.* a decision tree) we build realizes a partition of the domain $\mathcal{X}$ into disjoint subsets $X_1, X_2, ..., X_N$. Fix as $[\![\pi]\!]$ the function returning the truth value of a predicate $\pi$. Define

$$W_+^{j,l} = \sum_{(o,c_o)\in LS} w((o,c_o))[\![(o,c_o)\in X_j \wedge c_o = l]\!] \ ,$$

$$W_-^{j,l} = \sum_{(o,c_o)\in LS} w((o,c_o))[\![(o,c_o)\in X_j \wedge c_o \neq l]\!] \ .$$

In other words, $W_+^{j,l}$ represents the fraction of examples of class $l$ present in subset $X_j$, and $W_-^{j,l}$ represents the fraction of examples of classes $\neq l$ present in subset $X_j$. According to Schapire & Singer (1998), a weak learner should minimize the criterion:

$$Z = 2\sum_j \sum_l \sqrt{W_+^{j,l} W_-^{j,l}} \; . \quad (1)$$

In the case of a decision tree, the partition is that which is built at the leaves of the tree (Quinlan, 1994) ; in the case of a decision list, the partition is that which is built at each rule, to which we add the subset associated to the default class (Nock & Jappy, 1998). Suppose that we encode the decision tree in the form of a subset of monomials, by taking for each leaf the logical-$\wedge$ of all attributes from the root to the leaf. Measuring $Z$ over the tree's leaves is equivalent to measure $Z$ over the partition realized by the set of monomials. However, the monomials are disjoint from each other (each example satisfies exactly one monomial). Due to this property, only $t$ subsets can be realized with $t$ monomials, or equivalently with a tree having $t$ leaves.

Suppose that we generalize this observation by removing the disjointness condition over the monomials. Then a number of subsets of order $\mathcal{O}(2^t)$ is now possible with only $t$ monomials, and it appears that the number of realized partitions can be exponentially larger using decision committees than decision trees. However, the expected running time is not bigger when using decision committees, since the number of partitions is in fact bounded by the number of examples, $|LS|$. Thus, we may expect some reduction in the size of the formula we build when using decision committee, which is of interest to interpret the classifier obtained.

Application of this principle in WIDC is straightforward: a large decision committee is built by growing iteratively, in a top-down fashion, a current monomial. In this monomial, the literal added at the current step is the one which minimizes the current $Z$ criterion, over all possible addition of literals, and given that the new monomial does not exist already in the current decision committee (in order to prevent multiple additions of a single monomial). The $Z$ criterion is computed using the partition induced over $LS$ by the current set of monomials built (if two examples satisfy the same monomials, they belong to the same subset of the partition). When no further addition of a literal decreases the $Z$ value, a new monomial is created and initialized at $\emptyset$, and then is grown using the same principle. When no further creation of a monomial decreases the $Z$ value, the algorithm stops and returns the current, large decision committee with still empty vectors. In the following step, WIDC calculates these vectors. In a previous approach to building rule sets for problems with two classes (Cohen & Singer, 1999), an iterative growing-pruning algorithm is designed (SLIPPER). The rule-growing approach of SLIPPER is certainly close to what WIDC does for growing a DC since it optimizes a $Z$ criterion, yet a notable difference is that it does not compute $Z$ over a partition induced by a set of rules. Rather, the choice of SLIPPER is to grow at each step a *single* monomial, prune it, and then grow a second monomial, prune it, and so on until a final DNF-shaped formula is complete and returned. Notice that SLIPPER also modifies the weight of the examples, in accordance with Boosting's standards (Schapire & Singer, 1998).

### 5.2 Calculating Rule Vectors using Ranking Loss Boosting

Schapire & Singer (1998) have investigated classification problems where the aim of the procedure is not to provide an accurate class for some observation. Rather, the algorithm outputs a set of values (one for each class) and we expect the class of the observation to receive the largest value of all, thus being ranked higher than all others. This approach is particularly useful when a given example can belong to more than one class (multilabel problems), a case where we expect each of these classes to receive the greatest values compared to the classes the examples does not belong to.

The *ranking loss* represents informally the number of times the hypothesis fails to rank the class of an example higher than a class to which it does not belong. Before going further, we first generalize our classification setting, and replace the common notation $(o, c_o)$ for an example by the more general one $(o, \vec{c}_o)$. Here, $\vec{c}_o \in \{0,1\}^c$ is a vector giving, for each class, the membership to the class ("0" is no and "1" is yes) of the corresponding observation $o$. It is important to note that this setting is more general than the usual Bayesian setting, in which there can exist examples $(o, c_o)$ and $(o', c_{o'})$ (using the non-vector notation) for which $o = o'$ but $c_o \neq c_{o'}$. Ranking loss generalizes Bayes to the multilabel problems, and postulates that there can be some examples for which we cannot provide a single class at a time, even if *e.g.* any of the classes to which the example belongs are susceptible to appear independently later with the same observation.

Ranking loss Boosting replaces each example $(o, \vec{c}_o)$ by a set of $1_{\vec{c}_o} \times (c - 1_{\vec{c}_o})$ examples, where $1_{\vec{c}_o}$ denotes the Hamming weight of $\vec{c}_o$ (*i.e.* the number of classes to which the example belongs). Each of these new examples is denoted $(o, k, j)$, where $j$ and $k$ span all values in $\{0, 1, ..., c-1\}^2$. The distribution of the new examples is renormalized, so that $w((o,k,j)) = \frac{w((o,\vec{c}_o))}{1_{\vec{c}_o} \times (c - 1_{\vec{c}_o})}$ whenever $\vec{c}_o[j] = 1$ and $\vec{c}_o[k] = 0$, and 0 otherwise.

Take some monomial $t$ obtained from the large DC, and all examples satisfying it. We now work with this restricted subset of examples, while calculating the corresponding vector $\vec{v}$ of $t$. Schapire & Singer (1998) propose a cost function which we should minimize in order to minimize the ranking loss. This function is

$$Z = \sum_{o,j,k} w((o,k,j)) \times e^{-\frac{1}{2}\alpha(\vec{v}[j] - \vec{v}[k])} . \tag{2}$$

Here, $\alpha$ is a tunable parameter which, intuitively, represents the confidence in the choice of $\vec{v}$, and leverages its quality. The better $\vec{v}$ is at classifying examples, the larger is $|\alpha|$. In our case however, authorizing $\alpha \neq 1$ is equivalent to authorizing components for $\vec{v}$ in sets $\{-x, 0, x\}$ for arbitrary $x$. To really constrain the components of $\vec{v}$ in $\{-1, 0, +1\}$, we have chosen to optimize the criterion

$$Z = \sum_{o,j,k} w((o,k,j)) \times e^{-\frac{1}{2}(\vec{v}[j] - \vec{v}[k])} \tag{3}$$

(therefore forcing $\alpha = 1$). Schapire & Singer (1998) conjecture that finding the optimal vector minimizing $Z$ in eq. (2) (which is similar to an *oblivious* hypothesis according to their definitions), or $Z$ given a particular value of $\alpha$, is $NP$-Hard when $c$ is not fixed, and when the components of $\vec{v}$ are in the set $\{-1, +1\}$. The following section addresses directly the setting of Schapire & Singer (1998), and presents complexity-theoretic results showing that the minimization of $Z$ is actually polynomial, but highly complicated to achieve, all the more for what it is supposed to bring to the minimization of $Z$ in our setting. A striking result we also give, not related to the purpose of the paper, is that it is actually the maximization of $Z$ which is $NP$-Hard.

Then, we present the approximation algorithm we have built and implemented to optimize the computation of $\vec{v}$ in our setting (components of $\vec{v}$ in the set $\{-1, 0, +1\}$), along with its properties. While we feel that the ideas used to minimize $Z$ in the setting of Schapire & Singer (1998) can be adapted to our setting to provide an algorithm that is always optimal, our algorithm has the advantage to be simple, fast, and also optimal for numerous cases. In many other cases, we show that it is still asymptotically optimal as $c$ increases.

#### 5.2.1 Optimizing $Z$ in the Setting of Schapire & Singer (1998)

In the case where each component of $\vec{v}$ is restricted to the set $\{-1, +1\}$, Schapire & Singer (1998) give a way to choose $\alpha$ to minimize $Z$ for any possible choice of $\vec{v}$ (using our notation):

$$\alpha = \frac{1}{2}\log\left(\frac{W^+}{W^-}\right) , \tag{4}$$

with:

$$W^+ = \sum_{o,k,j} w((o,k,j))[\![\vec{v}[j] - \vec{v}[k] = 2]\!] , \tag{5}$$

$$W^- = \sum_{o,k,j} w((o,k,j))[\![\vec{v}[j] - \vec{v}[k] = -2]\!] . \tag{6}$$

Replacing this value of $\alpha$ in eq. (2), gives the following new expression for $Z$:

$$Z = W^0 + 2\sqrt{W^+W^-} , \tag{7}$$

with $W^0 = \sum_{o,k,j} w((o,k,j))[\![\vec{v}[j] - \vec{v}[k] = 0]\!]$. Schapire & Singer (1998) raise the problem of minimizing $Z$ as defined in equations (2) and (7). We now show that it is polynomial.

**Theorem 4** *Minimizing $Z$ as defined either in equations (2), (3) or (7) is polynomial when the components of $\vec{v}_i$ are restricted to the set $\{-1, +1\}$.*

**Proof:** See the Appendix. □

A rather striking result given the conjecture of Schapire & Singer (1998) is that it is the maximization of $Z$, and not its minimization, which is $NP$-Hard. While this is not the purpose of the present paper (we are interested in minimizing $Z$), we have chosen to give here a brief proof sketch of the result, which uses classical reductions from well-known $NP$-Hard problems.

**Theorem 5** *Maximizing $Z$ as defined either in equations (2), (3) or (7) is $NP$-Hard when the components of $\vec{v}_i$ are restricted to the set $\{-1, +1\}$.*

**Proof sketch:** See the Appendix. □

### 5.2.2 Optimizing $Z$ in our Setting

As previously argued in Theorem 4, minimizing $Z$ in the setting of Schapire & Singer (1998) can be done optimally, but at the expense of complex optimization procedures, with large complexities. One can wonder whether such procedures, to optimize only the computation of $\vec{v}$ (a small part of WIDC), are really well worth the adaptation to our setting, in which more values are authorized. We are now going to show that a much simpler combinatorial procedure, with comparatively very low complexity, can bring optimal results in fairly general situations. The most simple way to describe most of these situations is to make the following assumption on the examples:

($\mathcal{A}$) Each example used to compute $\vec{v}$ has only one "1" in its class vector.

A careful reading of assumption ($\mathcal{A}$) reveals that it implies that each example belongs to exactly one class, *but* it does not prevent an observation to be element of more than one class, as long as different examples sharing the same observation have different classes (the "1" of the class vectors is in different positions among these examples). Therefore, even if it does not integrate the most general features of the ranking loss setting, our assumption still authorizes to consider problems with non zero Bayes optimum. This is really interesting, as many commonly used datasets fall into the category of our assumption, as for example many datasets of the UCI repository of Machine Learning database (Blake et al., 1998). Finally, even if the assumption does not hold, we show that in many of the remaining (interesting) cases, our approximation algorithm is asymptotically optimal, that is, finds solutions closer to the minimal value of $Z$ as $c$ increases.

Suppose for now that ($\mathcal{A}$) holds. Our objective is to calculate the vector $\vec{v}$ of some monomial $t$. We use the shorthands $W_0^+, W_1^+, ..., W_{c-1}^+$ to denote the sum of weights of the examples satisfying $t$ and belonging respectively to classes $0, 1, ..., c-1$. We want to minimize $Z$ as proposed in eq. (3). Suppose without loss of generality that

$$W_0^+ \leq W_1^+ \quad \leq \quad ... \leq W_{c-1}^+ \ ,$$

otherwise, reorder the classes so that they verify this assertion. Given only three possible values for each component of $\vec{v}$, the testing of all $3^c$ possibilities for $\vec{v}$ is exponential and time-consuming. But we can propose a very fast approach. We have indeed

**Lemma 1** $\forall 1 \leq j < k \leq c, \vec{v}[j] \leq \vec{v}[k]$ .

**Proof:** See the Appendix. □

Thus, the optimal $\vec{v}$ does not belong to a set of cardinality $3^c$, but to a set of cardinality $\mathcal{O}(c^2)$. Our algorithm is then straightforward: simply explore this set of $\mathcal{O}(c^2)$ elements, and keep the vector having the lowest value of $Z$. Note that this combinatorial algorithm has the advantage to be adaptable to more general settings in which $l$ particular values are authorized for the components of $\vec{v}$, for any fixed $l$ not necessarily equal to 3. In that case, the complexity is larger, but limited to $\mathcal{O}(c^{l-1})$.

There are slightly more general settings in which our algorithm remains optimal, in particular when we can certify $\forall j, k$, $((W_j^+ > W_k^+) \Leftrightarrow (\forall i \neq j, k : W_{j,i} > W_{k,i})) \vee ((W_j^+ < W_k^+) \Leftrightarrow (\forall i \neq j, k : W_{j,i} > W_{k,i}))$. Here, $W_x^+$ denotes the sum of weights of the examples belonging at least to class $x$, and $W_{x,y}$ denotes the sum of weights of the examples belonging at least to class $x$, and *not* belonging at least to class $y$. This shows that even for some particular multilabel cases, our approximation algorithm can remain optimal. One can wonder if the optimality is preserved in the unrestricted multilabel framework. We now show that, if optimality is not preserved, we can still prove the quality of our algorithm for general multilabel cases, showing asymptotic optimality as $c$ increases.

Our approximation algorithm is run in the multilabel case by transforming the examples as follows: each example $(o, \vec{c}_o)$ for which $1_{\vec{c}_o} > 1$ is transformed into $1_{\vec{c}_o}$ examples, having the same description $o$, and only one "1" in their vector, in such a way that we span the $1_{\vec{c}_o} > 1$ "1" of the original example. Their weight is the one of the original example, divided by $1_{\vec{c}_o}$. We then run our algorithm on this new set of examples satisfying assumption $(\mathcal{A})$.

Now, suppose that for any example $(o, \vec{c}_o)$, we have $1_{\vec{c}_o} \leq k$ for some $k$. There are two interesting vectors we use. The first one is $\vec{v}^*$, the optimal vector (or an optimal vector) minimizing $Z$ over the original set of examples, the second one is $\vec{v}$, the vector we find minimizing $Z$ over the transformed set of examples. What we want is to estimate the quality of $\vec{v}$ with respect to the optimal value of $Z$ over the original set of examples, $Z(\vec{v}^*)$ using our notation. The following theorem gives an answer to this problem, by quantifying its convergence towards $Z(\vec{v}^*)$.

**Theorem 6** $Z(\vec{v}) < Z(\vec{v}^*)\left(1 + \left(\frac{e}{c-k}\right)\right)$.

**Proof:** See the Appendix. □

Therefore, in the set of all problems for which for some $\beta < 1$, $k \leq \beta c$, we obtain $Z(\vec{v}) = (1 + o(1))Z(\vec{v}^*)$, and our bound converges to the optimum as $c$ increases in this class of problems. By means of words, our simple approximation algorithm is quite efficient for problems with large number of classes. Note that using a slightly more involved proof, we could have reduced the constant "$e$" factor in Theorem 6 to the slightly smaller "$e - (1/e)$". Now, to fix the ideas, the following subsection displays the explicit (and simple) solution when there are only two classes.

#### 5.2.3 Explicit Solution in the Two-Classes Case

For the sake of simplicity, rename $W_0^+ = W^-$ and $W_1^+ = W^+$ representing the fraction of examples from the negative and positive class respectively, satisfying $t$. The rule to choose $\vec{v}$ is the following:

**Lemma 2** *The following table gives the rule to choose $\vec{v}$ :*

| *If* | *then we choose* |
|---|---|
| $\frac{W_+}{W_-} \geq e^{\frac{3}{2}}$ | $\vec{v} = (-1, +1)$ |
| $\sqrt{e} \leq \frac{W_+}{W_-} < e^{\frac{3}{2}}$ | $\vec{v} = (-1, 0)$ *or* $\vec{v} = (0, +1)$ |
| $\frac{1}{\sqrt{e}} \leq \frac{W_+}{W_-} < \sqrt{e}$ | $\vec{v} = (-1, -1)$ *or* $\vec{v} = (0, 0)$ *or* $\vec{v} = (+1, +1)$ |
| $\frac{1}{e^{\frac{3}{2}}} \leq \frac{W_+}{W_-} < \frac{1}{\sqrt{e}}$ | $\vec{v} = (0, -1)$ *or* $\vec{v} = (+1, 0)$ |
| $\frac{W_+}{W_-} < \frac{1}{e^{\frac{3}{2}}}$ | $\vec{v} = (+1, -1)$ |

**Proof:** See the Appendix. □

### 5.3 Pruning a DC

The algorithm is a single-pass algorithm: each rule is tested only once, from the first rule to the last one. For each possible rule, a criterion `Criterion`(.) returns "TRUE" or "FALSE" depending on whether the rule should be removed or not. There are two versions of this criterion. The first one, which we call "pessimistic", is based on conventional error minimization. The second one, called "optimistic", is derived from a previous work on pruning decision-trees (Kearns & Mansour, 1998).

#### 5.3.1 Pessimistic Pruning

Pessimistic pruning builds a sequence of DC from the initial one. At each step, we remove one rule, such that its removal brings the lowest error among all possible removals of rule in the current DC. Each time the error of the current DC is not greater than the lowest error found already, `Criterion`(.) returns true for all rules already tested for removal. This pruning returns the smallest DC having the lowest error of the sequence. This pruning is rather natural (and simple), and motivated by the fact that the induction of the large DC before pruning does not lead to a conventional error minimization. Such a property is rather seldom in "top-down and prune" induction algorithms. For example, common decision tree induction algorithms in this scheme incorporate very sophisticated pruning criteria (CART (Breiman et al., 1984), C4.5 (Quinlan, 1994)).

#### 5.3.2 Optimistic Pruning

Kearns & Mansour (1998) present a novel algorithm to prune decision trees, based on a test over locally observed errors. Its principle is simple: each internal node of a DT is tested only once in a bottom-up fashion, and we estimate the local error over the learning examples reaching this node, before and after the removal of the node. If the local error after removal is not greater than the local error before, plus a penalty term, then we remove the node and its subtree. The penalty term makes the pruning essentially optimistic, that is, we tend to overprune the decision tree. However, thanks to local uniform convergence results, and due to the fact that certain sub-classes of decision trees are reasonably large, Kearns & Mansour (1998) are able to prove that with high probability, the overpruning will not be too severe with respect to the optimal subtree of the initial DT. We refer the reader to their paper for further theoretical results, not needed here. The point is that by using the results of Kearns & Mansour (1998), we can obtain a similar test for DC. We emphasize that our bound might not enjoy the same theoretical properties as for decision trees, because of the cardinality reasons briefly outlined before. However, such a test is interesting since it may lead especially to very small and interpretable decision committees, with the obvious hope that their accuracy will not decrease too much. Furthermore, the paper of Kearns & Mansour (1998) does not contain experimental results. We think our criterion as a way to test heuristically the experimental feasibility of some of the results of Kearns & Mansour (1998). The principle of our criterion is exactly the same as the original test of Kearns & Mansour (1998) : "can we compare, when testing some rule $(t, \vec{v})$ and using the examples that satisfy the rule, the errors before and after removing the rule"? Let $\epsilon_{(t,\vec{v})}$ represent the error before removing the rule, on the local sample $LS_{(t,\vec{v})}$ satisfying monomial $t$. Denote $\epsilon_{\emptyset}$ as the error before removing $(t, \vec{v})$, still measured on the local sample $LS_{(t,\vec{v})}$. Then we define the heuristic "penalty" (proof omitted: it is a rough upperbound of Kearns & Mansour (1998), Lemma 1) :

$$\alpha'_{(t,\vec{v})} = \sqrt{\frac{(\mathrm{Set}((t,\vec{v})) + 2)\log(n) + \log 1/\delta}{|LS_{(t,\vec{v})}|}} . \quad (8)$$

$\mathrm{Set}((t,\vec{v}))$ denotes the maximum number of literals of all rules except $(t,\vec{v})$ in the current DC, that an arbitrary example could satisfy. The fast calculability of $\alpha'_{(t,\vec{v})}$ is obtained at the expense of a greater risk of overpruning, whose effects on some small datasets were experimentally dramatic for the accuracy. In our experiments, which contain very small datasets, we have chosen to tune a parameter limiting the effects of this combinatorial upperbound. More precisely, We have chosen to uniformly resample $LS$ into a larger subset of 5000 examples, when the initial $LS$ contained less than 5000 examples. By this, we artificially increase $|LS_{(t,\vec{v})}|$ and mimic for the small domains new domains with an identical larger size, with the additional benefits that reasonable comparisons may be made on pruning.

The value of `Criterion`$((t,\vec{v}))$ is therefore "TRUE" iff $\epsilon_{(t,\vec{v})} + \alpha'_{(t,\vec{v})} \geq \epsilon_{\emptyset}$.

## 6. Experiments

Following are three experimental sections, aimed at testing WIDC on three issues. The first presents extensive results on the tradeoff simplicity-accuracy obtained by WIDC, and compares the results with those obtained for state-of-the-art algorithms. The second goes on in depth analyzes for the mining/interpretability issue, and the third presents results on noise tolerance.

### 6.1 Tradeoff Simplicity-Accuracy

Experiments were carried out using three variants of WIDC: with optimistic pruning (o), with pessimistic pruning (p), and without pruning ($\emptyset$). Table 1 presents some results on various datasets, most of which were taken from the UCI repository of machine learning database (Blake et al., 1998). For each dataset, the eventual discretization of attributes was performed following previous recommendations and experimental setups (de Carvalho Gomes & Gascuel, 1994). The results were computed using a ten-fold stratified cross validation procedure (Quinlan, 1996). The least errors for WIDC are underlined for each domain. For the sake of comparisons, column "Others" points out various results for other algorithms, intended to help getting a general picture of what can be the performances of efficient approaches with different outputs (decision lists, trees, committees, etc.), in terms of errors (and, when applicable, sizes). Some of the most relevant results for WIDC are summarized in the scatterplots of Table 2.

The interpretation of Table 1 using only errors gives the advantage to WIDC with pessimistic pruning, all the more as WIDC(p) has the advantage of providing simpler formulas than WIDC($\emptyset$), and has a much simpler pruning stage than WIDC(o). Results also compare favorably to the "Other" results, building either DLs, DTs, or DCs. They are all the more interesting if we compare the errors in the light of the sizes obtained. For the "Echo" domain, WIDC with pessimistic pruning beats improved CN2 by two points, but the DC obtained contains roughly eight times fewer literals than CN2-POE's decision list. If we except "Vote0", on all other problems on which we dispose of CN2-POE's results, we outperform CN2-POE on both accuracy and size. Finally, on "Vote0", note that WIDC with optimistic pruning is slightly outperformed by CN2-POE by 2.51%, but the DC obtained is *fifteen* times smaller than the decision list of CN2-POE. If we dwell on the results of C4.5, similar conclusions can be brought: on 12 out of 13 datasets on which we ran C4.5, WIDC(p) finds smaller formulas, and still beats C4.5's accuracy on 9 of them. A quantitative comparison of $l_{DC}$ against the number of nodes of the DTs shows that on 4 datasets out of the 13 (Pole, Shuttle, TicTacToe, Australian), the DCs are more than 6 times smaller, while they only incur a loss in accuracy for 2 of them, and limited to 1.8%. For this latter problem (TicTacToe), a glimpse at Table 1 shows that the DCs, with less than 7 rules on average, keeps comparatively most of the information contained in DTs having more than a hundred leaves. On many problems where mining issues are crucial, such a size reduction would be well worth the (comparatively slight) loss in accuracy, because we keep a significant part of the information on very small classifiers, thus likely to be interpretable.

| Domain | WIDC(o) err% | $r_{DC}$ | $l_{DC}$ | WIDC(p) err% | $r_{DC}$ | $l_{DC}$ | WIDC($\emptyset$) err% | $r_{DC}$ | $l_{DC}$ | Other |
|---|---|---|---|---|---|---|---|---|---|---|
| Australian | 15.57 | 1.1 | 1.8 | 16.00 | 1.6 | 4.1 | 18.14 | 4.8 | 17.5 | $15.1_{39.0}$ **f** |
| Balance | 22.38 | 4.1 | 10.5 | 14.76 | 9.9 | 27.3 | 14.29 | 18.7 | 44.9 | $20.1_{86.0}$ **f** |
| Breast-W | 7.46 | 1.1 | 4.5 | 4.08 | 5.0 | 21.0 | 6.90 | 7.7 | 29.3 | $4.9_{21.8}$ **f** |
| Bupa | 36.57 | 3.2 | 12.4 | 37.14 | 4.3 | 16.6 | 37.14 | 7.7 | 28.4 | $37.3_{37.0}$ **f** |
| Echo | 32.14 | 1.8 | 3.9 | 27.86 | 4.7 | 11.1 | 31.42 | 24.6 | 38.8 | $32.3_{35.4}$ **a** |
| Glass2 | 21.76 | 1.5 | 4.7 | 21.17 | 1.7 | 5.4 | 26.47 | 4.3 | 12.5 | $26.3_{8.0}$ **f** |
| Heart-S | 24.07 | 3.1 | 8.9 | 19.48 | 8.5 | 31.4 | 21.85 | 12.5 | 40.8 | 21.5 **c** |
| Heart-C | 22.90 | 2.9 | 9.1 | 21.85 | 6.5 | 27.4 | 25.48 | 13.3 | 46.2 | $22.5_{52.0}$ **a** |
| Heart-H | 22.67 | 3.9 | 10.9 | 20.45 | 8.4 | 24.2 | 20.00 | 14.3 | 43.5 | $21.8_{60.3}$ **a** |
| Hepatitis | 20.59 | 3.4 | 8.7 | 19.24 | 7.0 | 17.0 | 15.29 | 11.4 | 26.7 | $19.2_{34.0}$ **a** |
| Horse | 15.26 | 1.7 | 3.6 | 15.57 | 3.8 | 10.4 | 20.26 | 12.5 | 31.7 | $15.7_{13.4}$ **f** |
| Iris | 5.33 | 1.9 | 4.6 | 5.33 | 2.9 | 7.1 | 20.67 | 3.7 | 7.9 | 8.5 **c** |
| Labor | 15.00 | 2.9 | 5.0 | 15.00 | 3.7 | 6.6 | 16.67 | 3.8 | 6.7 | $16.31_{6.8}$ **d** |
| LED7 | 31.09 | 6.9 | 8.4 | 24.82 | 16.2 | 21.3 | 24.73 | 19.0 | 25.4 | $25.73_{12.2}$ **d** |
| LEDeven | 13.17 | 2.7 | 6.1 | 12.43 | 3.8 | 9.2 | 24.63 | 9.9 | 21.9 | $13.00_{19.2}$ **f** |
| LEDeven2 | 30.00 | 4.1 | 16.4 | 23.15 | 7.1 | 26.1 | 21.70 | 24.4 | 83.8 | $23.1_{25.4}$ **f** |
| Lung | 42.50 | 1.3 | 3.8 | 42.50 | 2.6 | 7.1 | 42.50 | 2.7 | 7.2 | 46.6 **e** |
| Monk1 | 15.00 | 4.1 | 9.5 | 15.00 | 5.2 | 13.0 | 15.00 | 9.4 | 17.9 | $16.66_{5.0}$ **d** |
| Monk2 | 24.43 | 9.0 | 38.4 | 21.48 | 18.2 | 61.3 | 31.80 | 24.8 | 82.1 | $29.39_{18.0}$ **d** |
| Monk3 | 3.04 | 3.6 | 4.8 | 9.89 | 4.7 | 8.9 | 12.50 | 9.3 | 12.3 | $2.67_{2.0}$ **d** |
| Pima | 29.61 | 2.2 | 5.9 | 26.17 | 8.0 | 29.4 | 32.99 | 22.2 | 68.9 | 25.9 **c** |
| Pole | 36.67 | 1.5 | 4.1 | 33.52 | 4.2 | 12.7 | 37.64 | 24.0 | 65.8 | $35.5_{81.6}$ **f** |
| Shuttle | 3.27 | 1.0 | 2.0 | 3.27 | 1.0 | 2.0 | 4.51 | 2.0 | 4.0 | $1.7_{29.8}$ **f** |
| TicTacToe | 22.47 | 5.7 | 14.4 | 20.10 | 6.7 | 17.6 | 23.50 | 15.9 | 43.7 | $18.3_{130.9}$ **f** |
| Vehicle2 | 26.47 | 2.8 | 7.8 | 26.70 | 4.0 | 11.2 | 33.18 | 16.4 | 46.5 | $25.6_{43.0}$ **f** |
| Vote0 | 6.81 | 1.9 | 3.0 | 8.40 | 4.5 | 8.5 | 10.00 | 9.5 | 18.9 | $4.3_{49.6}$ **a** |
| Vote1 | 10.90 | 2.0 | 3.5 | 9.98 | 7.0 | 14.9 | 12.50 | 13.6 | 29.7 | $10.89_{6.4}$ **d** |
| Waveform | 30.49 | 4.8 | 8.2 | 23.47 | 7.5 | 17.3 | 20.24 | 40.1 | 65.0 | $33.5_{21.8}$ **b** |
| Wine | 10.00 | 3.0 | 6.2 | 9.47 | 3.7 | 8.1 | 7.89 | 4.2 | 8.9 | 22.8 **e** |
| XD6 | 16.73 | 5.2 | 14.4 | 17.50 | 6.2 | 17.1 | 22.69 | 19.8 | 52.0 | $21.2_{58.0}$ **f** |

Table 1: Experimental results using WIDC.
Conventions: $l_{DC}$ is the whole number of literals of a DC, $r_{DC}$ is its number of rules. For "Others", numbers are given on the form $\text{error}_{(\text{size})}$, where **a** is improved CN2 (CN2-POE) building DLs, size is the number of literals (Domingos, 1998). **b** is ICDL building DLs, notations follow **a** (Nock & Jappy, 1998). **c** is C4.5 (Franck & Witten, 1998). **d** is IDC building DCs, notations follow **a** (Nock & Gascuel, 1995). **e** is 1-Nearest Neighbor rule and **f** is C4.5 (pruned, default parameters) building DTs; the size of a tree is its whole number of nodes.

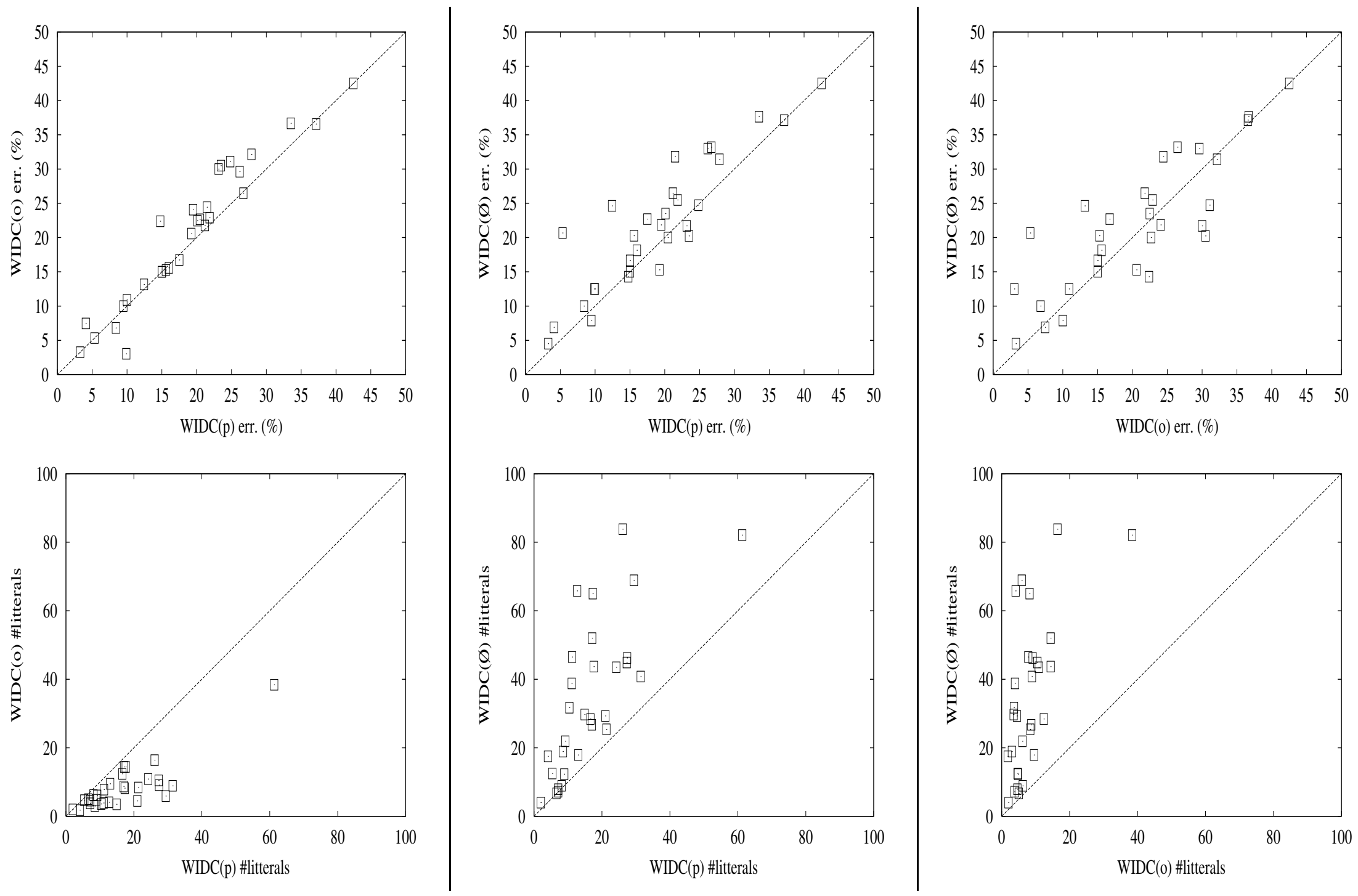

Table 2: Scatterplots summarizing some results of Table 1 for the three flavors of WIDC, in terms of error (first row) and size ($l_{DC}$, second row), on the thirty datasets. Each point above the $x = y$ line depicts a dataset for which the algorithm in abscissa performs better.

### 6.2 Interpretability Issues

In the XD6 domain, each example has 10 binary variables. The tenth is irrelevant in the strongest sense (John, Kohavi, & Pfleger, 1994). The target concept $f$ is a 3-DNF (a DNF with each monomial containing at most three literals) over the first nine variables: $(x_0 \wedge x_1 \wedge x_2) \vee (x_3 \wedge x_4 \wedge x_5) \vee (x_6 \wedge x_7 \wedge x_8)$. Such a formula is typically hard to encode using a small decision tree. In our experiments with WIDC(o) and WIDC(p), we have remarked that the target formula itself is almost always an element of the classifier built, and the irrelevant attribute is always absent. Figure 1 shows an example of DC which was obtained on a run of WIDC. Note that the concept returned is a 3-DC. Figure 2 depicts a part of a tree obtained on this domain with C4.5. While the tree appears to be quite large for the domain, note the presence of the irrelevant variable in the tree, which it contributes to enlarge while making it harder to mine. On many other domains, we observed persistent rules or subconcepts through the 10 cross-validation runs. Similarly to XD6, whenever we could mine the results with a sufficiently accurate knowledge of the domain, these patterns were most interesting. For example, the DCs obtained on the LEDeven domain contained most of the time a combination of two rules with one literal each, which represented a very accurate way to classify 9 out of the 10 possible classes. On the Vote0 and Vote1 domains, we also observed constant patterns, some of which are well known (Blake et al., 1998) to provide a very accurate classification for a tiny size. Even for Vote1 where classical studies often report errors over 12%, and almost never around 10% (Holte, 1993), we observed on most of the runs a DC containing an accurate rule with two literals only, with which WIDC(p) provided on average an error under 10%.

| | − | + |
|---|---|---|
| $x_0 \wedge x_1 \wedge x_2$ | -1 | 1 |
| $x_3 \wedge x_4 \wedge x_5$ | -1 | 1 |
| $x_6 \wedge x_7 \wedge x_8$ | -1 | 1 |
| $x_1 \wedge \overline{x}_5$ | 0 | -1 |
| $\overline{x}_0 \wedge x_2$ | 1 | -1 |
| $\overline{x}_1$ | 1 | -1 |
| default $\vec{D}$ | 0.963 | 0.037 |

Figure 1: A DC obtained on the XD6 domain with WIDC(p). The first three rules exactly encode the target concept, and the irrelevant variable is absent from the DC.

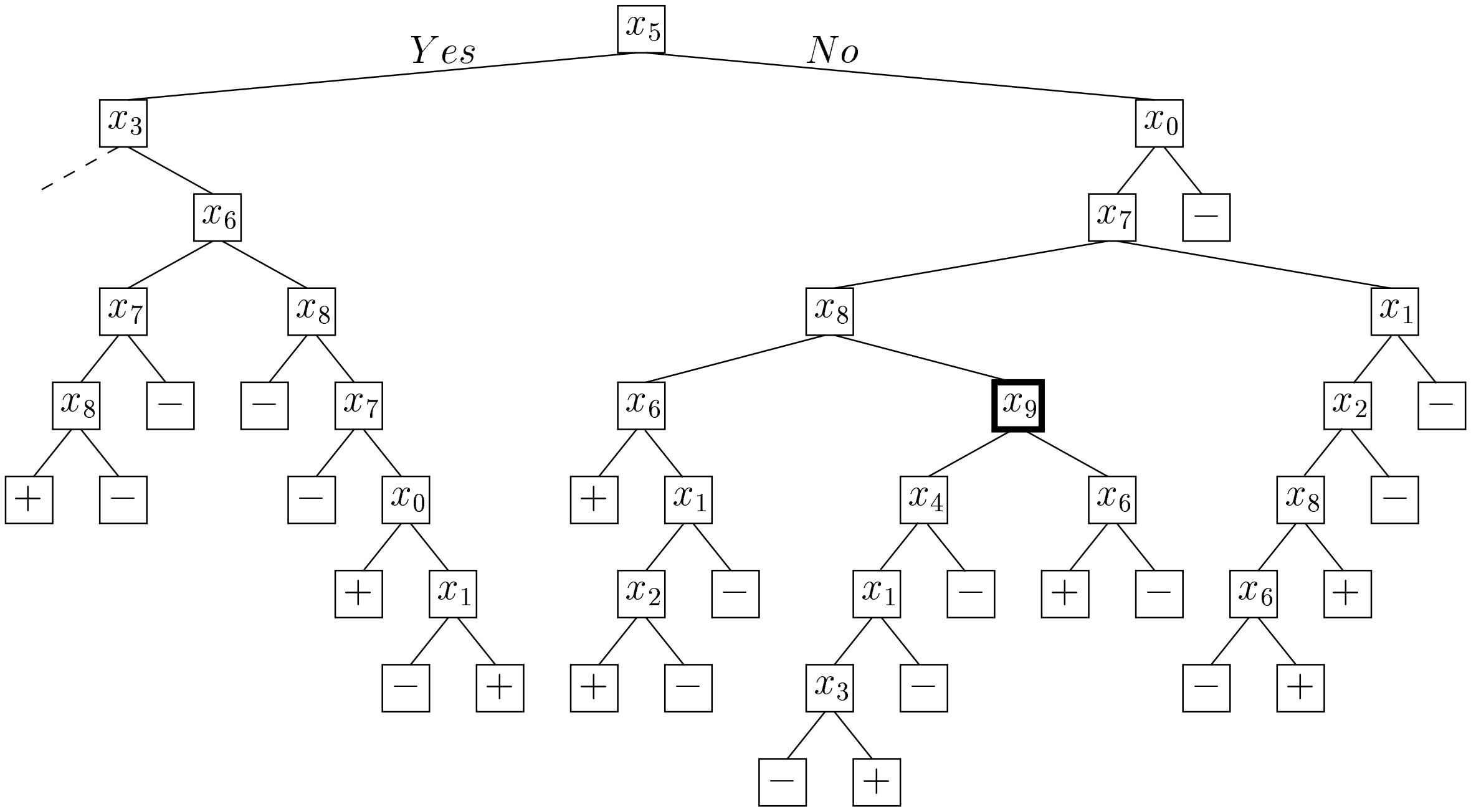

Figure 2: Part of a DT obtained on the XD6 domain with C4.5. Positive literals label the internal nodes. To classify an observation, the left edge of a node is followed when an observation contains ("*Yes*") the positive literal, and the right edge is followed otherwise (*i.e.* the literal is negative in the observation). The bold square is used to display the presence of the irrelevant variable in the tree. A naive conversion of this tree in rules for both classes generates 30 rules, for a total of 179 literals.

WIDC was also compared to C4.5 on a real world domain on which mining issues are as crucial as classification strength: agriculture. An experiment is being carried out in Martinique by the DDAF (Departmental Direction of Agriculture and Forest), to achieve better understanding of the behavior of farmers, in particular regarding their willingness to contract a CTE (Farming Territorial Contract). Usual farming contracts with either the state (France) or Europe did not contain commitments for the farmer to satisfy. In a CTE, each farmer commits to adapt and/or change his agricultural techniques or productions, to ensure sustainable development for local agriculture. In exchange for this, he receives the guarantee to obtain financial help for this contract, and to be trained to new agricultural techniques. Such a domain is a good test bed to evaluate a method on the basis of predictability and interpretability, because of the place of uncertainty in agriculture, and the fact that obtaining data can be a hard and long task : the DDAF has to be as accurate as possible in its predictions and interpretations, to manage as best as possible its relationships with farmers, and in the case of CTEs, to make the best promotion campaign for these new contracts. Agriculture is also very sensitive to a "showcase effect": provided even few representative farmers will have subscribed to the contracts, comparatively many others are likely to follow.

In this study, from the description of 52 variables for about 60 representative farmers satisfying the criteria to adhere to a CTE, the aim is to develop models for those who are actually willing to adhere, those not willing to adhere, and those currently uncertain. Variables are data on each agricultural exploitation (size, terrain nature, financial data, type of production, etc .), as well as more personal data on the farmers (education, family status, objectives, personal answer to a questionnaire, etc.). This represents a small dataset to mine, but, interestingly, the results obtained were different when processing it with C4.5 or WIDC(p).

| | adhere | ? | ¬adhere |
|---|---|---|---|
| $(No_ongoing_project) \wedge (No_education)$ | -1 | -1 | 1 |
| $(No_ongoing_project) \wedge (Lengthy_proc) \wedge (No_Wholesaler)$ | 1 | -1 | 1 |
| default $\vec{D}$ | 0.32 | 0.68 | 0 |

Figure 3: The DC obtained on the agricultural data (see text for the interpretation of the variables).

We ran both algorithms in a 10-fold stratified cross-validation experiment. WIDC(p) obtained a 2.8% average error. In 6 out of 10 runs, the same DC was induced. It is presented in Figure 3. Basically, this DC proves that predicting the "¬adhere" class is the easiest task, followed by the prediction of the "adhere" class. The "?" (uncertain farmers) is predicted only by the default vector. This seems rather natural: whereas the extreme behaviors tend to be clear to determine, the uncertainty is the hardest to predict.

C4.5 (default parameters) induced a DT which was almost the exact transcription of rule 1, a rule which says that farmers with no education (without any agricultural diploma or traineeships) and no ongoing project are not willing to adhere. This rule is mostly interesting because it proves that education is a strong factor determining the "¬adhere" answer. The DTs induced also contained one or two more literals separating the "adhere" and "?" classes (average error: 6.7%), but only few other things could be mined from the trees of C4.5, in the light of the problem addressed.

Rule 2 in Figure 3 did not have the equivalent in the DTs induced. What it says is interesting for the DDAF, because it brings the following conclusion: farmers without ongoing projects, and not selling their products only to a wholesaler, are on the knife edge for their membership (either in "adhere", or in "¬adhere"). Without going further into local agricultural considerations, this rule, for the DDAF Engineers, represents an accurate view of the farmers actually controlling their exploitation costs, being either for or against CTEs, and that education pushes towards the membership (combination of rules 1 and 2), probably because it allows them to see the future potential benefits of the contract, better than its current constraints.

### 6.3 Noise Handling

Noise handling is a crucial issue for boosting (Bauer & Kohavi, 1999; Opitz & Maclin, 1999), even considered (Bauer & Kohavi, 1999) as its potential main problem. Experimental studies show that substantial noise levels can alter the vote to the point that its accuracy is lower than that of a single of its classifier (Opitz & Maclin, 1999). Opitz & Maclin (1999) point out the reweighting scheme of the examples in boosting as being a potential reason for this behavior. Though we do not use any reweighting scheme, we have chosen for the sake of completeness to address the behavior of WIDC(p) against noise, and compare its results with perhaps the major induction algorithm with which we share the "top-down and prune" induction scheme: C4.5 (Quinlan, 1994). This study relies on the XD6 domain, in which we replace the original 10% class noise (Buntine & Niblett, 1992) by various increasing amounts of class noise ranging from 0% to 40% by steps of 2%, or various increasing amounts of attribute noise in the same range. The XD6 domain has the advantage that the target concept is known, and it has been addressed in a substantial amount of previous experimental works. We have simulated corresponding datasets of 512 examples each, for each noise level. Each such dataset was processed by WIDC(p) and C4.5, using a 10-fold-cross-validation procedure. Figure 4 depicts the results obtained for the errors and for the sizes of the classifiers. The size of a DC is its whole number of literals, and that of a DT is its number of internal nodes.

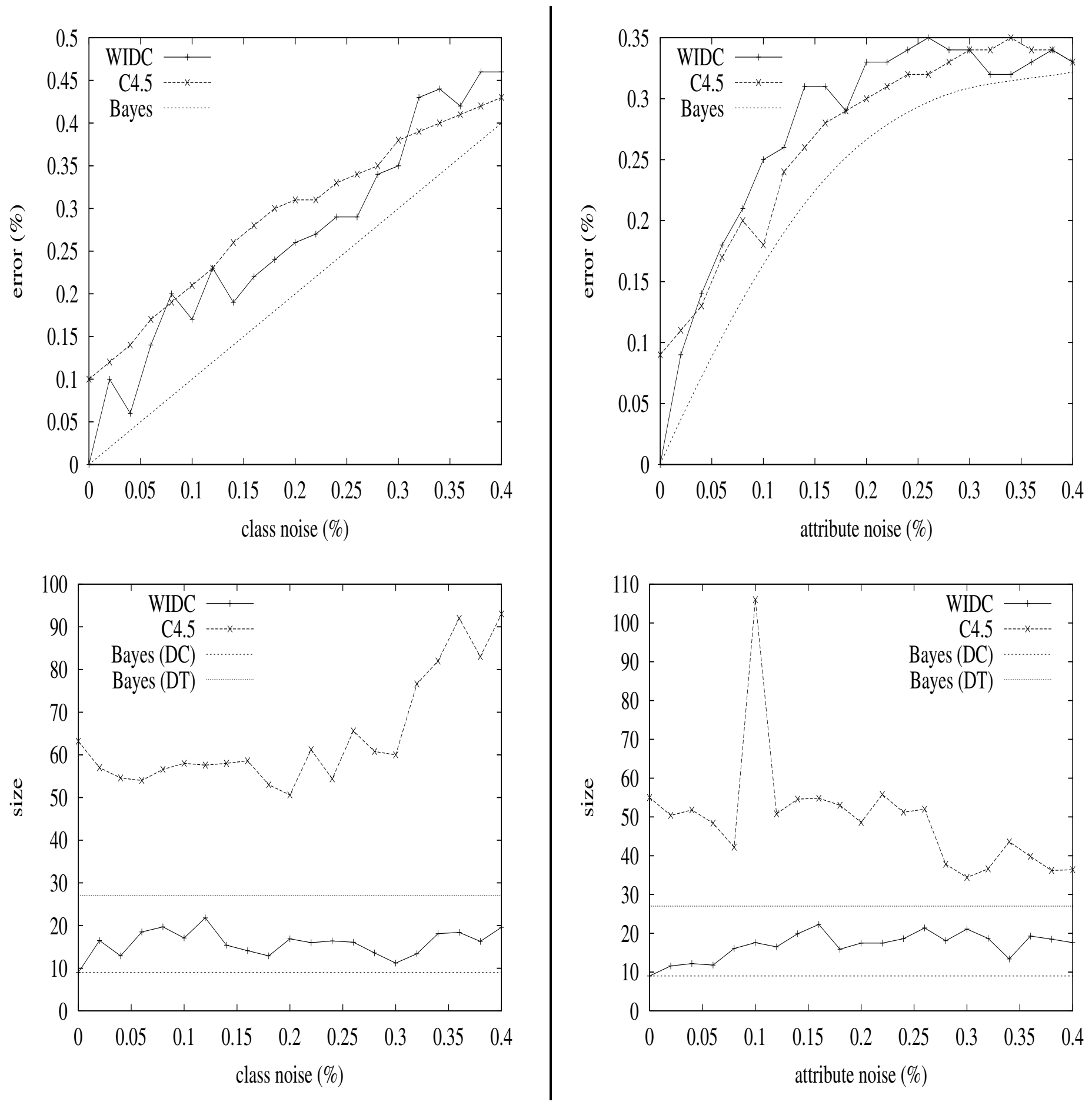

Figure 4: Plots of the errors (up) and sizes (down) of WIDC(p), C4.5 and Bayes rule against various class noise levels (left) and attribute noise levels (right).

While the resistance against noise seems to be relatively well distributed among WIDC(p) and C4.5 (WIDC(p) seems to perform better for class noise, while C4.5 seems to perform better for attribute noise), a phenomenon more interesting comes from the sizes of the formulas induced. First, the DCs have very small size fluctuations compared to the DTs : for class noises greater than 20%, the DTs have size increasing by a factor of 1.5-2. Second, note that the ratio between the number of nodes of the target DT, and the number of literals of the target DC is 3. For a majority of class or attribute noise levels, the ratio between the DTs build and the DCs built is $> 3$, with a pathologic case for 10% attribute noise, for which the ratio is $> 6$. These remarks, along with the fact that the DCs built have a very reasonable size when compared to that of the target DC for any type and level of noise, tend to show a good noise handling for WIDC(p). Apart from these considerations, glimpses at the DCs output by WIDC(p) show that even for large noise levels, it manages to find concepts syntactically close to the target DC. For example, one of the DCs output at 30% class noise is exactly the target DC; also, it is only for class noise $\geq 12\%$ (and attribute noise $\geq 16\%$) that some DCs found do not syntactically include the target DC anymore.

## 7. Conclusion

Recent advances in the study of voting classification algorithms have brought empirical and theoretical results clearly showing the discrimination power of ensemble classifiers. This paper addresses from a theoretical and empirical point of view the question of whether one might have to dispense with interpretability in order to keep classification strength. In order to cope with this problem, we have chosen to study a class of concept representations resembling multilinear discriminant polynomials, adequate for mining issues when dealing with voting procedures, which we define as Decision Committees. Our theoretical results show that striving for simplicity is, like for many other classes of concept representations, a hard computational problem when dealing with DC or other complex voting procedures, and proves the heuristic nature of other results trying to prune adaptive boosting. This paper proposes to adapt a previous scheme to build weak learners, successful for the induction of decision trees and decision lists, to the case of DC. This is an original approach if we refer to the state-of-the-art algorithms building complex votes procedures, usually working in the strong learning framework. Our algorithm, WIDC, relies on recents or new results about partition boosting, ranking loss boosting, and pruning. It comes with two flavors, one with optimistic pruning and one with pessimistic pruning. Both obtained experimentally good results on the simplicity-accuracy tradeoff, but whereas optimistic pruning clearly outperforms other algorithms in the light of the size of the formulas obtained, pessimistic pruning tends to achieve a more reasonable tradeoff, with high accuracies obtained on small formulas. This is all the more interesting as pessimistic pruning is based on a natural and simple pruning procedure.

## 8. Acknowledgments

Thanks are due to DDAF Martinique, ENESAD (Etablissement National d'Enseignement Supérieur Agronomique de Dijon) and Lise Jean-Louis for having provided the agricultural data, for stimulating discussions around our results, and for having authorized the publication of some of the results obtained. Thanks to Ahmed Ainouche for having pointed out the interest in minimizing submodular functions. Finally, the author wishes to thank Pedro Domingos and the reviewers for their valuable suggestions.

# Appendix A

## Proof of Theorem 1

Since the hardness results of Theorems 1 and 3 are stated for the two-classes case, we shall use the notation $\Delta_{(i)} = \vec{v}_{(i)}[1] - \vec{v}_{(i)}[0]$ for some arbitrary rule $(t_{(i)}, \vec{v}_{(i)})$, where $\vec{v}_{(i)}[0]$ is the value for class "-" and $\vec{v}_{(i)}[1]$ is the value for class "+". A positive value for $\Delta_{(i)}$ means that $t_{(i)}$ is in favor of class "+" whereas a negative value gives a $t_{(i)}$ in favor of class "-". Value 0 for $\Delta_{(i)}$ gives a $t_{(i)}$ neutral with respect to the classes. We use a reduction from the $NP$-Hard problem "Minimum Cover" (Garey & Johnson, 1979):

- **Name** : "Minimum Cover".
- **Instance** : A collection $C$ of subsets of a finite set $S$. A positive integer $K$, $K \leq |C|$.
- **Question** : Does $C$ contain a cover of size at most $K$, that is, a subset $C' \subseteq C$ with $|C'| \leq K$, such that any element of $S$ belongs to at least one member of $C'$ ?

The reduction is constructed as follows : from a "Minimum Cover" instance we build a learning sample $LS$ such that if there exists a cover of size $|C'| \leq K$ of $S$, then there exists a decision committee with $|C'|$ literals consistent with $LS$, and, reciprocally, if there exists a decision committee with $k$ literals consistent with $LS$, then there exists a cover of size $k$ of $S$. Hence, finding the smallest decision committee consistent with $LS$ is equivalent to finding the smallest $K$ for which there exists a solution to "Minimum Cover", and this is intractable if $P \neq NP$.

Let $c_j$ denote the $j^{th}$ element of $C$, and $s_j$ the $j^{th}$ element of $S$. We define a set of $|C|$ Boolean variables in one to one correspondence with the elements of $C$, which we use to describe the examples of $LS$. The corresponding set of literals is denoted $\{x_1, \overline{x}_1, x_2, \overline{x}_2, ..., x_{|C|}, \overline{x}_{|C|}\}$. The sample $LS$ contains two disjoint subsets : the set of positive examples $LS^+$, and the set of negative ones $LS^-$. $LS^+$ contains $|S|$ examples, denoted by $e_1^+, e_2^+, ..., e_{|S|}^+$. We construct each positive example so that it encodes the membership of the corresponding element of $S$ in the elements of $C$. More precisely,

$$\forall 1 \leq i \leq |S|, e_i^+ = \left(\bigwedge_{j:s_i \in c_j} x_j\right) \wedge \left(\bigwedge_{j:s_i \notin c_j} \overline{x}_j\right) . \tag{9}$$

$LS^-$ contains a single negative example, defined by:

$$e^- = \bigwedge_{j=1}^{j=|C|} \overline{x}_j . \tag{10}$$

• Suppose there exists a cover $C'$ of $S$ satisfying $|C'| \leq K$. We create a decision committee consisting of $K$ monomials, each with one literal only and associated to a positive $\Delta$. Each monomial codes one of the sets in $C'$. The default class is "-". This decision committee is consistent with the examples of $LS^+ \cup LS^-$, otherwise some element of $S$ would not be covered. If there are only two values authorized for the vectors and they are $\leq 0$, we simply create a DC consisting of one monomial with negative literals associated to a negative $\Delta$ (the value for the negative class is greater than the one of the positive class); each of the negative literals codes one of the sets in $C'$. The default class is "+".
• Suppose now that there exists a decision committee $f$ with at most $k$ literals consistent with $LS$. Denote $t_1, t_2, ..., t_{|f|}$ each monomial of $f$, in no specific order, and $\Delta_1, \Delta_2, ..., \Delta_{|f|}$ their associated values for $\Delta$. The monomials of $f$ can belong to three types of subsets of monomials:

- monotonous monomials (without negative literals),
- monomials containing only negative literals,
- monomials containing positive and negative literals.

| | monomial $t_i$ | $\Delta_i$ |
|---|---|---|
| I | $M$ | $> 0$ |
| II | $M$ | $< 0$ |
| III | $N$ | $> 0$ |
| IV | $N$ | $< 0$ |
| V | $MN$ | $> 0$ |
| VI | $MN$ | $< 0$ |

Figure 5: The six possible cases of rules.

Let us call respectively $M$, $N$, $MN$ these three classes. Given that each monomial of $f$ can be associated to a positive or a negative $\Delta$, there exists on the whole six classes of rules, presented in Figure 5.

Any monomial of $f$ containing at least one positive literal can only be satisfied by positive examples. Therefore, if there exists rules belonging to class II or VI, we can remove them without losing consistency. Furthermore, since $e^-$ contains only negative literals, if we remove their negative literals from all rules belonging to class V (making them go to class I), we do not lose consistency. As a consequence, we can suppose without loss of generality that all rules of $f$ are in class I, III, or IV.

We now treat independently two cases, depending on whether the default class of $f$ is "+" or "-".

1. The default class is "-". Any positive example satisfies therefore a monomial in $f$. There can exist two types of positive examples: those satisfying at least one rule of class I, and those not satisfying any class I rule (therefore satisfying at least one rule of class III). $e^-$ satisfies all class III and IV rules. Therefore,

$$\sum_{(t_i, \vec{v}_i) \in f \cap (\text{ class III} \cup \text{class IV })} \Delta_i \leq 0 \ . \quad (11)$$

   This shows that, if a positive example not satisfying any class I rules would satisfy all class IV rules, then it would be misclassified, which is impossible by the consistency hypothesis. This gives an important property, namely that any positive example not satisfying any class I rule cannot satisfy all class IV rules. Let us call **P** this property in what follows. We now show how to build a valid solution to "Minimum Cover" with at most $k$ elements. For any positive example $e_i^+$,

- if $e_i^+$ satisfies at least one class I rule, choose in $C$ a subset of $S$ corresponding to a positive literal of some satisfied class I rule. This subset contains $e_i^+$.
- if $e_i^+$ does not satisfy any class I rule, there exists from **P** some class IV rule which is not satisfied. Among all negative literals of a class IV rule which is not satisfied by $e_i^+$, choose one which is positive in $e_i^+$ (causing it not to satisfy the rule), and then choose the corresponding element of $C$. This subset of $S$ contains $e_i^+$.

Iterating the above procedure for all positive examples, we obtain a cover of $S$ consisting of at most $k$ subsets of $S$.

2. The default class is "+". $e^-$ satisfies all class III and IV rules. Therefore,

$$\sum_{(t_i,\vec{v}_i)\in f\cap(\text{ class III } \cup \text{ class IV })} \Delta_i < 0 \ . \qquad (12)$$

Even if the inequality is now strict, it gives the same procedure for efficiently building the solution to "Minimum Cover" with at most $k$ elements, by using the same argument as in the preceeding case.

This ends the proof of Theorem 1.

### Proof of Theorem 3

We use a reduction from the $NP$-Hard problem "2-NM-Colorability" (Kearns et al., 1987):

- **Name** : "2-NM-Colorability".
- **Instance** : A finite set $S = \{s_1, s_2, ..., s_{|S|}\}$ and a collection of constraints over $S$, $C = \{c_1, c_2, ..., c_{|C|}\}$, such that $\forall i \in \{1, 2, ..., |C|\}, c_i \subseteq S$.
- **Question** : Does there exist a 2-NM-Coloration of the elements of $S$, *i.e.* a function $\chi : S \rightarrow \{1, 2\}$ such that

$$(\forall i \in \{1, 2, ..., |C|\}), (\exists s_k, s_l \in c_i) : \chi(s_k) \quad \neq \quad \chi(s_l) \ ?$$

The reduction is constructed as follows : from a "2-NM-Colorability" instance, we build a learning sample $LS$ such that if there exists a 2-NM-Coloration of the elements of $S$, then there exists a decision committee with two rules consistent with $LS$, and, reciprocally, if there exists a decision committee with two rules consistent with $LS$, then there exists a 2-NM-Coloration of the elements of $S$. Furthermore, there never exists a decision committee with only one rule consistent with $LS$. Hence, finding the decision committee with the smallest number of rules consistent with $LS$ is at least as hard as solving "2-NM-Colorability", and this is intractable if $P \neq NP$.

Let $c_j$ denote the $j^{th}$ element of $C$, and $s_j$ the $j^{th}$ element of $S$. We define a set of $|S|$ Boolean variables in one to one correspondence with the elements of $S$, which we use to describe the examples of $LS$. The corresponding set of literals is denoted $\{x_1, \overline{x}_1, x_2, \overline{x}_2, ..., x_{|S|}, \overline{x}_{|S|}\}$. Our reduction is made in the two-classes framework. The sample $LS$ contains two disjoint subsets : the set of positive examples $LS^+$, and the set of negative ones $LS^-$. $LS^+$ contains $|S|$ examples, denoted by $e_1^+, e_2^+, ..., e_{|S|}^+$. We construct each positive example so that it represents an element of $S$. More precisely,

$$\forall 1 \leq i \leq |S|, e_i^+ \;=\; \overline{x}_i \wedge \bigwedge_{j=1, j\neq i}^{j=|S|} \overline{x}_j \ . \tag{13}$$

$LS^-$ contains $|C|$ examples, denoted by $e_1^+, e_2^+, ..., e_{|C|}^+$. We construct each negative example so that it encodes each of the constraints of $C$. More precisely:

$$\forall 1 \leq i \leq |C|, e_i^- \;=\; \left(\bigwedge_{j: s_j \in c_i} \overline{x}_j\right) \wedge \left(\bigwedge_{j: s_j \not\in c_i} x_j\right) \ . \tag{14}$$

Without loss of generality, we make four assumptions on the instance of "2-NM-Colorability" due to the fact that it is not trivial:

1. There does not exist some element of $S$ present in all constraints. In this case indeed, the trivial coloration consists in giving to one of such elements one color, and the other color to all other elements of $S$.

2. $\forall (i, j, k, l) \in \{1, 2, ..., |S|\}^4$ with $i \neq j$ and $k \neq l$,
$$\exists o \in \{1, 2, ..., |C|\}, \{s_i, s_j\} \not\subseteq c_o \quad \wedge \quad \{s_k, s_l\} \not\subseteq c_o \ . \tag{15}$$
Otherwise indeed, there would exist $(i, j, k, l) \in \{1, 2, ..., |S|\}^4$ with $i \neq j$ and $k \neq l$ such that
$$\forall o \in \{1, 2, ..., |C|\}, \{s_i, s_j\} \subseteq c_o \quad \vee \quad \{s_k, s_l\} \subseteq c_o \ , \tag{16}$$
and in that case, a trivial solution to "2-NM-Colorability" would consist in giving to $s_i$ one color and to $s_j$ the other one, and to $s_k$ one color and to $s_l$ the other one.

3. Each element of $S$ belongs to at least one constraint in $C$. Otherwise, it can be removed.

4. Each constraint contains at least two elements from $S$. Otherwise it can be removed.

• Suppose there exists a solution to "2-NM-Colorability". We build the DNF with two monomials of (Kearns et al., 1987) consistent with the examples. Then, we build two rules by associating the two monomials to some (arbitrary) positive value. The default class is "-". This leads to a decision committee with two rules consistent with $LS$.
• Suppose that there exists a decision committee $f$ with at most two rules consistent with $LS$. We now show that there exists a valid 2-NM-Coloration of the elements of $S$. We first show three lemmas which shall be used later on. Then, we show that the decision committee is actually equivalent to a DNF with two monomials consistent with $LS$. We conclude by using previous results (Kearns et al., 1987) on how to transform this DNF into a valid 2-NM-Coloration of the elements of $S$.

**Lemma 3** *If a monomial is not satisfied by any positive example,*

- *either it contains at least two negative literals, or*
- *it is the monomial containing all positive literals:*

$$\bigwedge_{j=1}^{j=|S|} x_j \ .$$

(Proof straightforward).

**Lemma 4** *If a monomial is satisfied by all positive examples, it is empty.*

(Indeed, for any variable, there exist two positive examples having the corresponding positive literal, and the corresponding negative literal).

**Lemma 5** *$f$ contains exactly two rules.*

**Proof:** Suppose that $f$ contains one rule, whose monomial is called $t_1$. If the default class is "-", all positive examples satisfy $t_1$, which is impossible by Lemma 4: the monomial would be empty, and $f$ could not be consistent. If the default class is "+", the negative examples are classified by $t_1$ and therefore $\Delta_1 < 0$. Thus, no positive example satisfies $t_1$. From Lemma 3, either $t_1 = \bigwedge_{j=1}^{j=|S|} x_j$ and no negative example can satisfy it (impossible), or $t_1$ contains at least two negative literals, and the constraints all have in common two elements of $S$. Thus, the instance of "2-NM-Colorability" is trivial, which is impossible. This ends the proof of Lemma 5. □

We now show that the default class of $f$ is "-". For the sake of simplicity, we write the two monomials of $f$ by $t_1$ and $t_2$. The default class is denoted $\beta \in \{$"-", "+"$\}$. Making the assumption that $\beta =$"+" implies that all negative examples must satisfy at least one monomial in $f$.

- Suppose that $\Delta_1 < 0$ and $\Delta_2 < 0$. Then, no positive example can satisfy either $t_1$ or $t_2$. From the two possibilities of Lemma 3, only the first one is valid ($\bigwedge_{j=1}^{j=|S|} x_j$ cannot be satisfied by any negative example). Thus, $t_1$ and $t_2$ contain each at least two negative literals:

$$\{\overline{x}_i, \overline{x}_j\} \subseteq t_1 \ , \qquad (17)$$
$$\{\overline{x}_k, \overline{x}_l\} \subseteq t_2 \ . \qquad (18)$$

  We are in the second case of triviality of the instance of "2-NM-Colorability", since making the assumption that $f$ is consistent implies:

$$\exists o \in \{1, 2, ..., |C|\}, \{s_i, s_j\} \not\subseteq c_o \quad \wedge \quad \{s_k, s_l\} \not\subseteq c_o \ . \qquad (19)$$

- Suppose that $\Delta_1 < 0$ and $\Delta_2 > 0$. All negative examples must satisfy $t_1$. $t_1$ is forced to be monotonous since otherwise (given that $\beta =$"+") all negative examples would share a common negative literal, thus all constraints would share a common element of $S$, and the instance of "2-NM-Colorability" would be trivial. $t_2$ being satisfied by at least one positive example (otherwise, $f$ would be equivalent to a single-rule decision committee, and we fall in the contradiction of Lemma 5), it contains at most one negative literal. If it contains exactly one negative literal, it is satisfied by exactly one positive example, and we can replace it by the monotonous monomial with $|S|-1$ positive literals (we leave empty the position of the initial negative literal). Consequently, similarly for $t_1$, we can suppose that $t_2$ is monotonous. We distinguish two cases.

- If $|\Delta_1| > |\Delta_2|$, no positive example can satisfy $t_1$. By fact 3, $t_1 = \bigwedge_{j=1}^{j=|S|} x_j$, and no negative example can satisfy it, a contradiction ($f$ cannot be consistent).
- If $|\Delta_1| \leq |\Delta_2|$. $t_2$ cannot be empty; therefore it contains a certain number of positive literals. Each positive example satisfying $t_2$ must also satisfy $t_1$, since otherwise $f$ is not consistent; Since $t_1$ and $t_2$ are monotonous, $t_2$ is a generalization of $t_1$, and any example satisfying $t_1$ (in particular, the negative examples) must satisfy $t_2$, a contradiction.

Therefore $\beta =$"-". This forces all positive examples to satisfy at least one monomial of $f$. Recall that $f$ contains two monomials. Suppose that $\Delta_1 > 0$ and $\Delta_2 < 0$. It comes $t_1 = \emptyset$ (Lemma 4). All negative examples must satisfy $t_2$, and we also have $|\Delta_1| \leq |\Delta_2|$. No positive example can satisfy $t_2$, and Lemma 3 gives either $t_1 = \bigwedge_{j=1}^{j=|S|} x_j$ (satisfied by no example, impossible) or $t_2$ contains at least two negative literals, whose corresponding elements of $S$ are shared by all constraints, and we obtain again that the instance of "2-NM-Colorability" is trivial.

Therefore, $\Delta_1 > 0$ and $\Delta_2 > 0$, and each monomial is satisfied by at least one positive example. $f$ is thus equivalent to a DNF with the same two monomials, and we can use a previous solution (Kearns et al., 1987) to build a valid 2-NM-Coloration. First, we can suppose that $f$ is again monotonous (Kearns et al., 1987). Then, since each positive example satisfies at least one monomial ($\beta =$"-"), then for all variable, there exists a monomial which does not contain the corresponding positive literal. The 2-Coloration is then

$$\forall i \in \{1, 2, ..., |S|\}, \chi(s_i) = \min_{j \in \{1,2\}} \{j : x_i \not\in t_j\} \ . \tag{20}$$

Could this be invalid ? That would mean that there exists a constraint $c_i$ such that $\forall s_j \in c_i, \chi(s_j) = K = cst$. This would mean that the corresponding negative example satisfies $t_K$, a contradiction (Kearns et al., 1987). This ends the proof of Theorem 3.

**Proof of Theorem 4**

Define the function $f : 2^{\{0,1,\ldots,c-1\}} \rightarrow I\!\!R$ such that

$$\forall A \subseteq \{0, 1, ..., c-1\}, f[A] = \sum_{o,k,j} w((o, k, j)) q_A(j, k) \ , \tag{21}$$

with

$$q_A(j,k) = e^{\alpha}[\![j \in A \wedge k \not\in A]\!] + e^{-\alpha}[\![j \not\in A \wedge k \in A]\!] + [\![(j \in A \wedge k \in A) \vee (j \not\in A \wedge k \not\in A)]\!] \ .$$

| $k \in$ | $l \in$ | coefficient of $w((o,k,j))$ in $f[A \cup B] + f[A \cap B]$ | $f[A] + f[B]$ |
|---|---|---|---|
| $S_1$ | $S_1$ | $2$ | $2$ |
| $S_1$ | $S_2$ | $e^{-\alpha}+1$ | $e^{-\alpha}+1$ |
| $S_1$ | $S_3$ | $e^{-\alpha}+1$ | $e^{-\alpha}+1$ |
| $S_1$ | $S_4$ | $2e^{-\alpha}$ | $2e^{-\alpha}$ |
| $S_2$ | $S_1$ | $e^{\alpha}+1$ | $e^{\alpha}+1$ |
| $S_2$ | $S_2$ | $2$ | $2$ |
| $S_2$ | $S_3$ | $2$ | $e^{\alpha}+e^{-\alpha}$ |
| $S_2$ | $S_4$ | $1+e^{-\alpha}$ | $1+e^{-\alpha}$ |
| $S_3$ | $S_1$ | $e^{\alpha}+1$ | $e^{\alpha}+1$ |
| $S_3$ | $S_2$ | $2$ | $e^{\alpha}+e^{-\alpha}$ |
| $S_3$ | $S_3$ | $2$ | $2$ |
| $S_3$ | $S_4$ | $1+e^{-\alpha}$ | $1+e^{-\alpha}$ |
| $S_4$ | $S_1$ | $2e^{\alpha}$ | $2e^{\alpha}$ |
| $S_4$ | $S_2$ | $1+e^{\alpha}$ | $1+e^{\alpha}$ |
| $S_4$ | $S_3$ | $1+e^{\alpha}$ | $1+e^{\alpha}$ |
| $S_4$ | $S_4$ | $2$ | $2$ |

Table 3: Possible coefficients of $w((o,k,j))$. We have fixed for short $S_1 = \{0,1,...,c-1\}\backslash(A \cup B)$, $S_2 = A\backslash B$, $S_3 = B\backslash A$, $S_4 = A \cap B$.

Note that $f$ generalizes the three expressions of $Z$ in equations (2), (3), and (7) with adequate values for $\alpha$. Now, we check that $f$ satisfies the submodular inequality:

$$f[A \cup B] + f[A \cap B] \quad \leq \quad f[A] + f[B] \ , \tag{22}$$

for all subsets $A, B \subseteq \{0,1,...,c-1\}$. The key is to examine the coefficient of each $w((o,k,j))$, for each set $\{0,1,...,c-1\}\backslash(A \cup B)$, $A\backslash B$, $B\backslash A$, $A \cap B$ to which $j$ or $k$ can belong. Table 3 presents these coefficients. We get from Table 3 :

$$\begin{aligned} & f[A \cup B] + f[A \cap B] - (f[A] + f[B]) = \\ & \quad (2 - e^{\alpha} - e^{-\alpha}) \sum_{o,k,j} w((o,k,j))[\![(j \in A\backslash B \wedge k \in B\backslash A) \vee (j \in B\backslash A \wedge k \in A\backslash B)]\!] \ . \end{aligned}$$

This last quantity is $\leq 0$ for any possible choice of $\alpha$. Therefore, minimizing $Z$ in any of its three forms of eq. (2), (3), and (7) boils down to minimizing $f$ on the submodular system $(\{0,1,...,c-1\}, f)$ (with the adequate values of $\alpha$). This problem admits polynomial-time solving algorithms (Grötschel, Lovàsz, & Schrijver, 1981; Queyranne, 1998). What is much interesting is that the algorithms known are highly complicated and time consuming for the general minimization of $f$ (Queyranne, 1998). However, when using the value of $\alpha$ as in eq. (4) and $Z$ as in eq. (7), the corresponding function $f$ becomes submodular symmetric ($f[A] = f[\{0,1,...,c-1\}\backslash A]$). As such, more efficient (and simpler) algorithms exist to minimize $f$. For example, there exists a powerful combinatorial algorithm working in $\mathcal{O}(c^5)$ (Queyranne, 1998). Note that this is still a very large complexity.

**Proofsketch of Theorem 5**

The reduction is made from the $NP$-Hard problem 3SAT5 (Feige, 1996). This is the classical 3SAT problem (Garey & Johnson, 1979), but each variable appears in exactly 5 clauses. Using a well-known reduction (Garey & Johnson, 1979), page 55, with an additional simple gadget, we can make a reduction from 3SAT5 to vertex cover (thus, independent set), obtaining a graph $G$ in which all vertices have degree either 5, or 0, and for which the largest independent set (for satisfiable instances of 3SAT5) has size $|V|/2$, where $|V|$ is the number of vertices of $G$. From this particular graph, we build a simple reduction to our problem of maximizing $Z$. Note that since we are searching for an oblivious hypothesis, the observations are not important (we can suppose that all examples have the same observation). That is why the reduction only builds class vectors (over $|V|$ classes), encoding the class membership of any of these identical observations. The idea is that the classes are in one-to-one mapping with the vertices, and there are two sets of class vectors built from $G$:

- a set with $|V|$ vectors, encoding the vertices of $G$. Each one is a class vector with only one "1" corresponding to the vertex, and the remaining components are zeroes. Each of the corresponding examples have weight $W_v$.
- a set with $|E|$ vectors, where $|E|$ is the number of edges of $G$. Each one encodes an edge, and therefore contains two "1" (and the remaining are zeroes) corresponding to the two vertices of the edge. Each of the corresponding examples have weight $W_e$.

Consider formulas (2), (3) for example. They are the sum of the contribution to $Z$ of the examples having weight $W_v$, and the examples having weight $W_e$. In these cases, we can rewrite $Z$ using the generic expression:

$$Z = Z_v + Z_e \ , \quad (23)$$

$$Z_v = W_v \left(e^{-\alpha}k(|V|-k) + k(k-1) + (|V|-k)(|V|-k-1) + e^{\alpha}k(|V|-k)\right) \ , \quad (24)$$

$$Z_e = W_e \left(e^{-\alpha}(|V|-k)(2C+U) + e^{\alpha}k(2M+U)\right) \ . \quad (25)$$

Here, $C$ is the number of edges having their two vertices in the set corresponding to the $+1$ values in $\vec{v}_i$, $M$ is the number of edges having their two vertices in the set corresponding to the $-1$ values in $\vec{v}_i$, and $U$ is the number of edges having one of their vertices in the $+1$ set, and the other one in the $-1$ set. $k$ is the number of $+1$ values in $\vec{v}_i$.

Suppose that $W_v \gg W_e$ (*e.g.* $W_v > |V|^3 W_e$). Then the maximization of $Z$ is the maximization of $Z_v$, followed by the maximization of $Z_e$. $Z_v$ admits a maximum for $k = |V|/2$, and with this value for $k$, it can be shown that maximizing $Z_e$ boils down to maximize $2M + U$, that is, the (weighted) number of edges not falling entirely into the set corresponding to the $+1$ values; whenever the 3SAT5 instance is satisfiable (and using the particular degrees of the vertices), this set corresponds to the largest independent set of $G$.

**Proof of Lemma 1**

The proof of this lemma is quite straightforward, but we give it for completeness. $Z$ can be rewritten as

$$Z = \sum_{j \neq k} Z_{j,k} \ , \quad (26)$$

with

$$\forall j \neq k, Z_{j,k} = \frac{W_j^+}{c-1} e^{-\frac{1}{2}\Delta(j,k)} + \frac{W_k^+}{c-1} e^{\frac{1}{2}\Delta(j,k)} , \quad (27)$$

where $\Delta(j,k) = \vec{v}[j] - \vec{v}[k]$. Suppose for contradiction that for some $j < k$, $\Delta = \Delta(j,k) > 0$. We simply permute the two values $\vec{v}[j]$ and $\vec{v}[k]$, and we show that the new value of $Z$ after, $Z_{\mathbf{a}}$, is not greater than $Z$ before permuting, $Z_{\mathbf{b}}$. The difference between $Z_{\mathbf{a}}$ and $Z_{\mathbf{b}}$ can be easily decomposed using the notation $Z_{(i,j)\mathbf{b}}$ $(i, j \in \{0, 1, ..., c-1\}, i \neq j)$ as the value of $Z_{i,j}$ (eq. (27)) in $Z_{\mathbf{b}}$, and $Z_{(i,j)\mathbf{a}}$ $(i, j \in \{0, 1, ..., c-1\}, i \neq j)$ as the value of $Z_{i,j}$ (eq. (27)) in $Z_{\mathbf{a}}$. We also define:

$$\forall i, j, k \in \{0, 1, ..., c-1\}, i \neq j \neq k, Z_{(i,j,k)\mathbf{a}} = Z_{(i,k),\mathbf{a}} + Z_{(k,i),\mathbf{a}} + Z_{(j,k),\mathbf{a}} + Z_{(k,j),\mathbf{a}} . \quad (28)$$

We define in the same way $Z_{(i,j,k)\mathbf{b}}$. We obtain

$$Z_{\mathbf{a}} - Z_{\mathbf{b}} = \left(Z_{(j,k)\mathbf{a}} - Z_{(j,k)\mathbf{b}}\right) + \sum_{i=1, i \notin \{j,k\}}^{c} \left(Z_{(j,k,i)\mathbf{a}} - Z_{(j,k,i)\mathbf{b}}\right) . \quad (29)$$

Proving that $Z_{\mathbf{a}} - Z_{\mathbf{b}} \leq 0$ can be obtained as follows. First,

$$Z_{(j,k)\mathbf{a}} - Z_{(j,k)\mathbf{b}} = \frac{\left(W_k^+ - W_j^+\right) e^{-\frac{1}{2}\Delta}}{c-1} \left(1 - e^{\Delta}\right) \leq 0 .$$

We also have $\forall i \in \{1, 2, ..., c\} \backslash \{j, k\}$:

$$\begin{aligned} Z_{(j,k,i)\mathbf{a}} - Z_{(j,k,i)\mathbf{b}} &= \frac{2W_j^+}{c-1} e^{\frac{1}{2}\Delta_{i,k}} + \frac{2W_k^+}{c-1} e^{\frac{1}{2}\Delta_{i,j}} - \frac{2W_j^+}{c-1} e^{\frac{1}{2}\Delta_{i,j}} - \frac{2W_k^+}{c-1} e^{\frac{1}{2}\Delta_{i,k}} \\ &= \frac{2\left(W_k^+ - W_j^+\right)}{c-1} \left(e^{\frac{1}{2}\Delta_{i,j}} - e^{\frac{1}{2}\Delta_{i,k}}\right) \\ &= \frac{2\left(W_k^+ - W_j^+\right) e^{\frac{1}{2}\Delta_{i,j}}}{c-1} \left(1 - e^{\Delta}\right) \leq 0 . \end{aligned}$$

Here we have use the fact that $\Delta_{i,k} = \Delta_{i,j} + \Delta$. This shows that $Z_{\mathbf{a}} - Z_{\mathbf{b}} \leq 0$, and ends the proof of Lemma 1.

### Proof of Theorem 6

To avoid confusion, we call $Z'$ the value of $Z$ computed over the transformed set of examples, and $U(\vec{u})$ for $U \in \{Z, Z'\}$ and $\vec{u} \in \{\vec{v}^*, \vec{v}\}$ as the value of criterion $U$ using vector $\vec{u}$. It is simple to obtain a "sufficient" bound to check the theorem. We have

$$Z(\vec{v}) = Z'(\vec{v}) - \sum_{\substack{S \subseteq V \\ |S| > 1}} \frac{W_S}{|S|} \left( \sum_{i \in S} e^{\frac{1}{2}\vec{v}[i]} \times \sum_{j \in S \backslash \{i\}} e^{-\frac{1}{2}\vec{v}[j]} \right) , \quad (30)$$

$$Z'(\vec{v}^*) = Z(\vec{v}^*) + \sum_{\substack{S \subseteq V \\ |S| > 1}} \frac{W_S}{|S|} \left( \sum_{i \in S} e^{\frac{1}{2}\vec{v}^*[i]} \times \sum_{j \in S \backslash \{i\}} e^{-\frac{1}{2}\vec{v}^*[j]} \right) . \quad (31)$$

Here, $W_S$ is the sum of weights of the examples in the original set, whose vectors have "1" matching the elements of $S$. Note that $Z'(\vec{v}) \leq Z'(\vec{v}^*)$, since our algorithm is optimal, and we obtain

$$\begin{aligned} &Z(\vec{v}) \\ &\leq \ Z(\vec{v}^*) + \sum_{\substack{S \subseteq V \\ |S| > 1}} \frac{W_S}{|S|} \left( \sum_{i \in S} \left[ e^{\frac{1}{2}\vec{v}^*[i]} \times \sum_{j \in S \backslash \{i\}} e^{-\frac{1}{2}\vec{v}^*[j]} - e^{\frac{1}{2}\vec{v}[i]} \times \sum_{j \in S \backslash \{i\}} e^{-\frac{1}{2}\vec{v}[j]} \right] \right) . \end{aligned}$$

By taking only the positive part of the right-hand side, and remarking that

- $\forall S \subseteq V, |S| > 1, \forall i \in S$, $\sum_{j \in S \backslash \{i\}} e^{-\frac{1}{2}\vec{v}^*[j]} \leq e \left( \frac{|S|-1}{c-|S|} \right) \sum_{j \in V \backslash S} e^{-\frac{1}{2}\vec{v}^*[j]}$ (the right sum is $\geq (c - |S|)e^{-\frac{1}{2}}$ and the left one is $\leq (|S| - 1)\sqrt{e}$),
- the coefficient of $W_S$ in $Z^*$ is $\rho_S = \sum_{i \in S} e^{\frac{1}{2}\vec{v}^*[i]} \times \sum_{j \in V \backslash S} e^{-\frac{1}{2}\vec{v}^*[j]}$,

we get

$$\begin{aligned} Z(\vec{v}) \ &\leq \ Z(\vec{v}^*) + \sum_{\substack{S \subseteq V \\ |S| > 1}} e\frac{(|S| - 1)}{|S|(c - |S|)} W_S \rho_S \\ &< \ Z(\vec{v}^*) + e\frac{(k-1)}{k(c-k)} \sum_{\substack{S \subseteq V \\ |S| > 1}} W_S \rho_S \\ &< \ Z(\vec{v}^*) \left( 1 + \left( \frac{e}{c-k} \right) \right) \ , \end{aligned}$$

as claimed.

## Proof of Lemma 2

$Z$ becomes in that case

$$Z \ = \ W_+ e^{-\frac{1}{2}\Delta} + W_- e^{\frac{1}{2}\Delta} \ , \qquad (32)$$

where $\Delta = \vec{v}[1] - \vec{v}[0]$. There are five different values for $\Delta$, giving rise to nine different $\vec{v}$:

$$\begin{aligned} \Delta = \ &+2 \ \Rightarrow \vec{v} = (-1, +1) \ , \\ \Delta = \ &+1 \ \Rightarrow \vec{v} = (-1, 0) \vee \vec{v} = (0, +1) \ , \\ \Delta = \ &\ \ 0 \ \Rightarrow \vec{v} = (-1, -1) \vee \vec{v} = (0, 0) \vee \vec{v} = (+1, +1) \ , \\ \Delta = \ &-1 \ \Rightarrow \vec{v} = (0, -1) \vee \vec{v} = (+1, 0) \ , \\ \Delta = \ &-2 \ \Rightarrow \vec{v} = (+1, -1) \ . \end{aligned}$$

Fix $\Delta = k$ where $k \in \{-2,-1,0,1,2\}$. $\forall k \in \{-1,0,1,2\}$, the value $\Delta = k$ should be preferred to the value $\Delta = k-1$ iff the corresponding $Z$ is smaller, that is :

$$W_+ \times e^{-\frac{k}{2}} + W_- \times e^{\frac{k}{2}} \quad < \quad W_+ \times e^{-\frac{k-1}{2}} + W_- \times e^{\frac{k-1}{2}} \ . \tag{33}$$

Rearranging terms gives $W_- < W_+ \times \frac{1}{e^{k-\frac{1}{2}}}$. This leads to the rule of the lemma.